%% file: main.tex
\documentclass[runningheads]{llncs}

\usepackage[T5]{fontenc}
\usepackage{textcomp}
\usepackage{etoolbox}
\usepackage{caption} 
\usepackage{makecell} 
\usepackage[protrusion=false]{microtype}
\usepackage{mathptmx} 
\usepackage{booktabs}
\usepackage{lmodern}
\usepackage{graphicx}
\usepackage{multirow}
\usepackage{lscape}
\usepackage{longtable}
\usepackage{xcolor}
\usepackage{multirow}
\usepackage{pgf-pie}   
\usepackage{xcolor}
\usepackage{tablefootnote}
\usepackage{tabularx}
\usepackage{adjustbox}
\usepackage{float}
\usepackage{caption}
\usepackage{subcaption}
\usepackage{threeparttable}
\usepackage{flushend}
\usepackage{longtable}
\usepackage{tabularray}
\usepackage{multirow}
\usepackage{lscape}
\usepackage{dblfloatfix}
\usepackage{lipsum}
\usepackage{wrapfig}
\usepackage{enumitem}
\usepackage{soul, soulutf8}

\usepackage{helvet}  
\usepackage{courier}  
\usepackage{xcolor} 
\usepackage{pifont} 
\definecolor{lightblue}{rgb}{0.7, 0.85, 1.0}
\newcommand{\cmark}{\textcolor{green!60!black}{\ding{51}}} 
\newcommand{\xmark}{\textcolor{red}{\ding{55}}} 
\usepackage{arydshln}
\renewcommand{\figurename}{\textbf{Figure}} 
\renewcommand{\tablename}{\textbf{Table}}  
\usepackage[utf8]{inputenc}
\usepackage{microtype}
\usepackage{multirow}
\usepackage{lscape}
\usepackage{longtable}
\usepackage{xcolor}
\usepackage{multirow}
\usepackage{pgf-pie}   
\usepackage{tablefootnote}
\usepackage{tabularx}
\usepackage{adjustbox}
\usepackage{float}
\usepackage{graphicx}
\usepackage{subcaption}
\usepackage{threeparttable}
\usepackage{flushend}
\usepackage{longtable}
\usepackage{tabularray}
\usepackage{multirow}
\usepackage{wrapfig}
\usepackage{enumitem}
\usepackage{xcolor}
\usepackage{soul, soulutf8}
\usepackage{booktabs}
\usepackage{multirow}
\usepackage{graphicx}
\usepackage{caption}
\usepackage{booktabs}
\usepackage{microtype}
\usepackage{multirow}
\usepackage{lscape}
\usepackage{longtable}
\usepackage{url}
\usepackage{amsmath}
\usepackage{algorithm}
\usepackage{algpseudocode}
\usepackage{tabularray}
\UseTblrLibrary{diagbox}
\usepackage{inconsolata}
\usepackage{float}  
\usepackage{booktabs}
\usepackage{multirow}
\usepackage{graphicx}
\usepackage{caption}
\usepackage{booktabs}
\usepackage{microtype}
\usepackage{multirow}
\usepackage{lscape}
\usepackage{longtable}
\usepackage{url}
\usepackage{amsmath}
\usepackage{algorithm}
\usepackage{algpseudocode}
\usepackage[hidelinks]{hyperref}
\usepackage{cite}

\hypersetup{
    colorlinks=true,
    linkcolor=black,
    citecolor=blue,
    urlcolor=blue,
}
\usepackage{tabularray}
\renewcommand{\figurename}{\textbf{Figure}} 
\renewcommand{\tablename}{\textbf{Table}}  
\definecolor{darkred}{RGB}{180,0,0} 

\begin{document}







\title{ViMRHP: A Vietnamese Benchmark Dataset for Multimodal Review Helpfulness Prediction via \\ Human-AI Collaborative Annotation}

\titlerunning{ViMHRP via Human-AI Collaborative Annotation}

\author{Truc Mai-Thanh Nguyen\inst{1,2} \and
        Dat Minh Nguyen\inst{1,2} \and 
        Son T. Luu\inst{1,2} \and  \\
        Kiet Van Nguyen\inst{1,2}\thanks{Corresponding author: \href{mailto:kietnv@uit.edu.vn}{kietnv@uit.edu.vn}}}

\authorrunning{Truc Nguyen et al.}

\institute{
            Faculty of Information Science and Engineering, \\ 
            University of Information Technology, Ho Chi Minh City, Vietnam \and
            Vietnam National University, Ho Chi Minh City, Vietnam \\
\email{\{21522721,21521937\}@gm.uit.edu.vn}, \email{\{sonlt,kietnv\}@uit.edu.vn}}
\maketitle


\begin{abstract}
Multimodal Review Helpfulness Prediction (MRHP) is an essential task in recommender systems, particularly in E-commerce platforms. Determining the helpfulness of user-generated reviews enhances user experience and improves consumer decision-making. However, existing datasets focus predominantly on English and Indonesian, resulting in a lack of linguistic diversity, especially for low-resource languages such as Vietnamese. In this paper, we introduce \textbf{ViMRHP} (\underline{\textbf{Vi}}etnamese \underline{\textbf{M}}ultimodal \underline{\textbf{R}}eview \underline{\textbf{H}}elpfulness \underline{\textbf{P}}rediction), a large-scale benchmark dataset for MRHP task in Vietnamese. This dataset covers four domains, including 2K products with 46K reviews. Meanwhile, a large-scale dataset requires considerable time and cost. To optimize the annotation process, we leverage AI to assist annotators in constructing the ViMRHP dataset. With AI assistance, annotation time is reduced (90–120 seconds/task $\rightarrow$ 20–40 seconds/task) while maintaining data quality and lowering overall costs by approximately 65\%. However, AI-generated annotations still have limitations in complex annotation tasks, which we further examine through a detailed performance analysis. In our experiment on ViMRHP, we evaluate baseline models on human-verified and AI-generated annotations to assess their quality differences. The ViMRHP dataset is publicly available at \textcolor{blue}{\footnote{\url{https://github.com/trng28/ViMRHP}}}.

\end{abstract}


\input{Sections/1_Introduction}
\input{Sections/2_RelatedWork}
\input{Sections/3_AnnotationProcess}
\input{Sections/4_ViMRHP}

\input{Sections/5_Experiments}

\input{Sections/6_Results}
\input{Sections/7_Conclusion}


\begin{credits}
\subsubsection{\ackname} This research was supported by The VNUHCM - University of Information Technology's Scientific Research Support Fund.
\end{credits}
\appendix
\input{Sections/Appendix}

\clearpage
\bibliographystyle{splncs03_unsrt_custom}
\bibliography{ref}

\end{document}

%% file: Sections/1_Introduction.tex
\section{Introduction}


Review Helpfulness Prediction (RHP) has become a crucial research topic in E-commerce due to its role in assessing the helpfulness of user-generated reviews and their impact on consumer purchasing decisions. However, the large volume of reviews poses significant challenges in identifying helpful reviews. Previous methods primarily relied on semantic features, argument mining, and classification tasks \cite{yang2015semantic,chen2022argument,diaz2018modeling}, while recent multimodal RHP (MRHP) approaches \cite{liu2021multi,han2022sancl,nguyen2022adaptive,xu2024personalized} integrate text and images to enable a more comprehensive assessment of review helpfulness and enhance predictive performance, focus on ranking reviews based on their helpfulness by using both product information and user-generated reviews.

\begin{table}[!htpb]
\centering
    \renewcommand{\arraystretch}{1.1} 

\begin{tabular}{p{6cm} p{6cm}} 
\toprule
\textbf{Product Information} & \\
Set 2 sữa rửa mặt Good morning COSRX độ ph thấp dạng gel chiết xuất trà xanh - 150ml/tuýp & 
\vspace{0.005cm}
\includegraphics[width=0.12\linewidth]{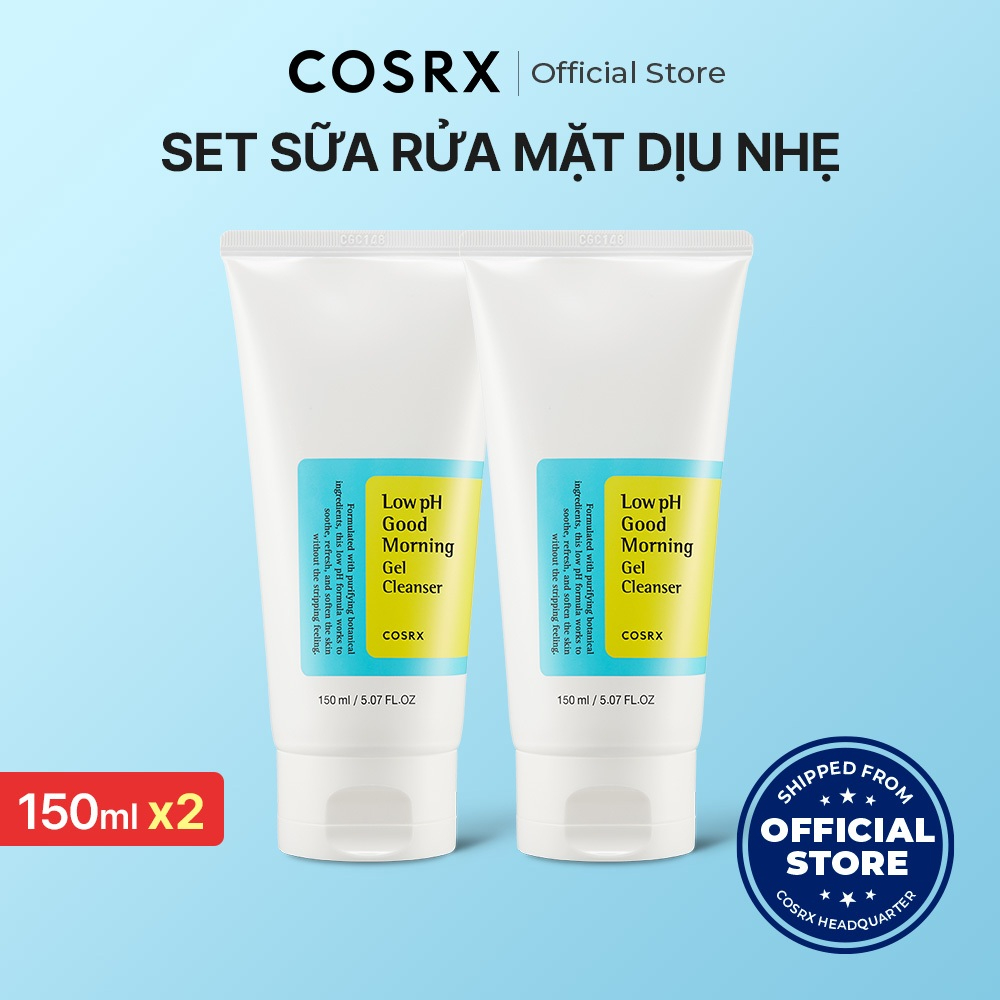}\hspace{0.02cm}
\includegraphics[width=0.12\linewidth]{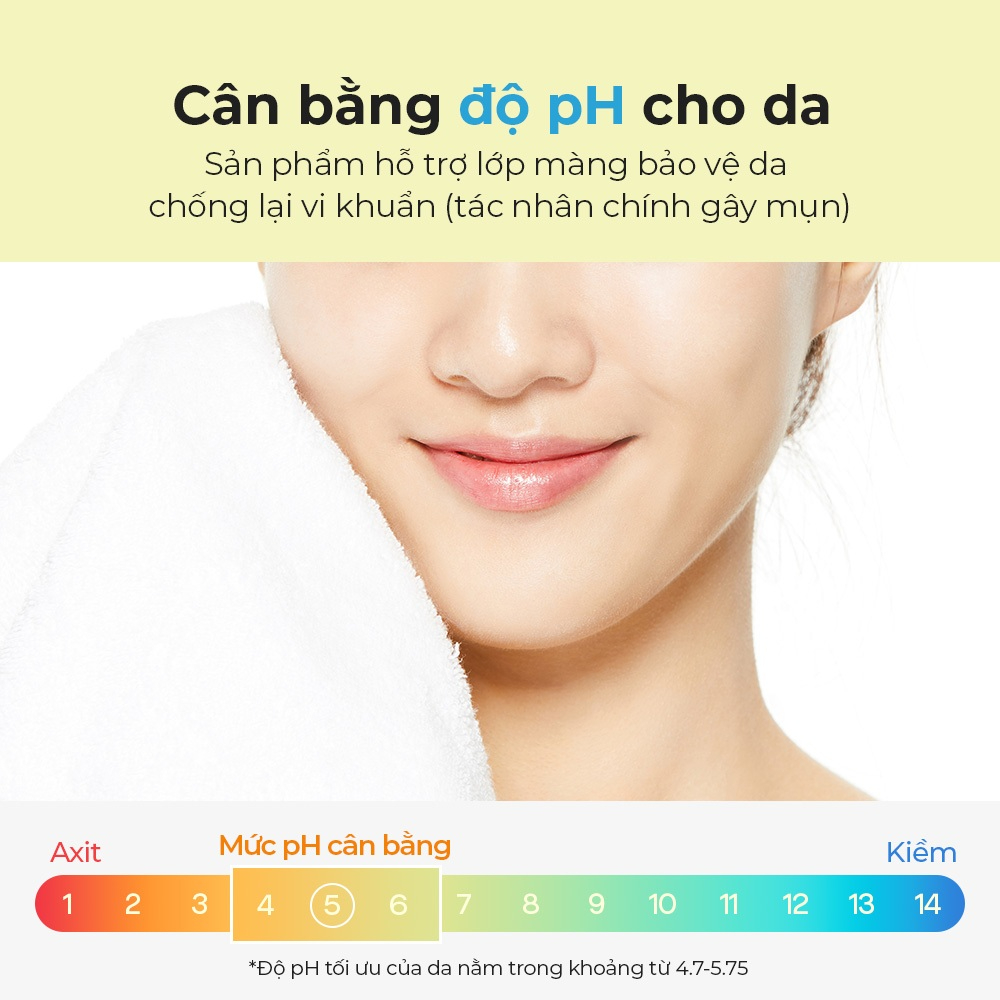}\hspace{0.02cm}
\includegraphics[width=0.12\linewidth]{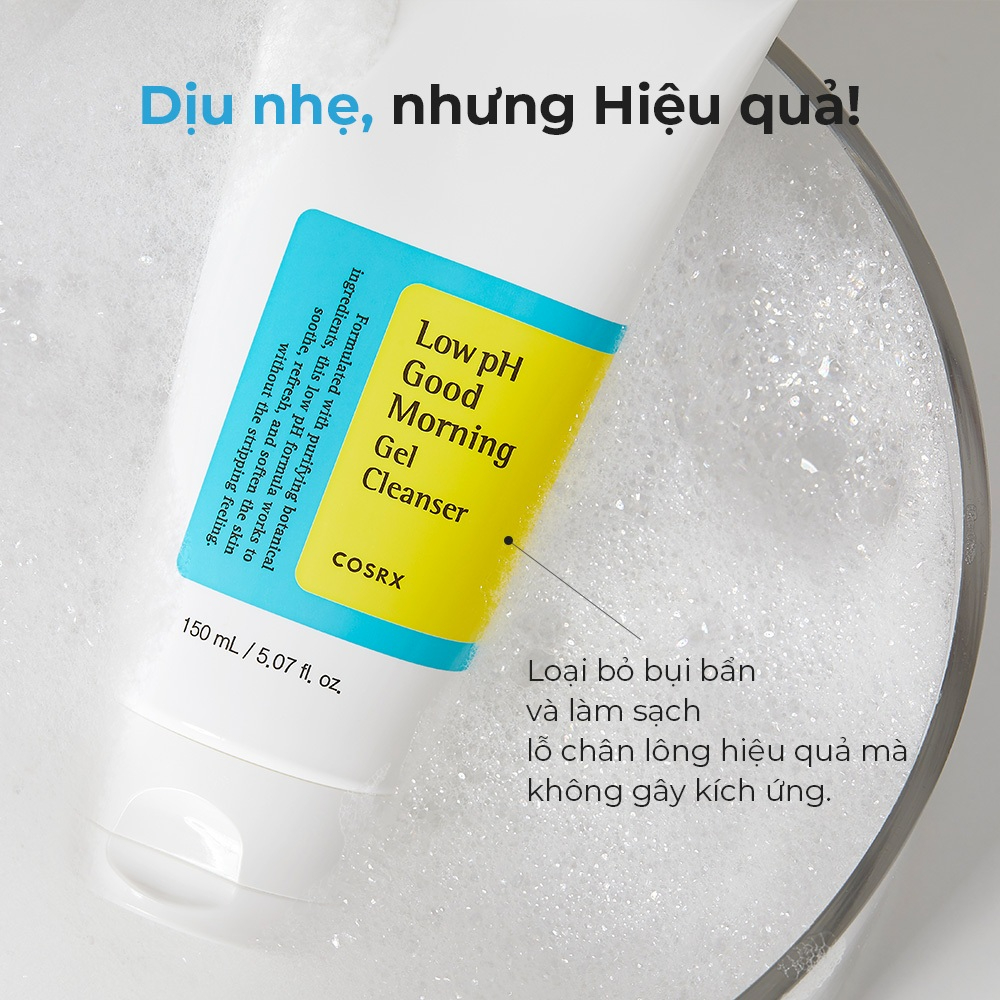}\hspace{0.02cm}
\includegraphics[width=0.12\linewidth]{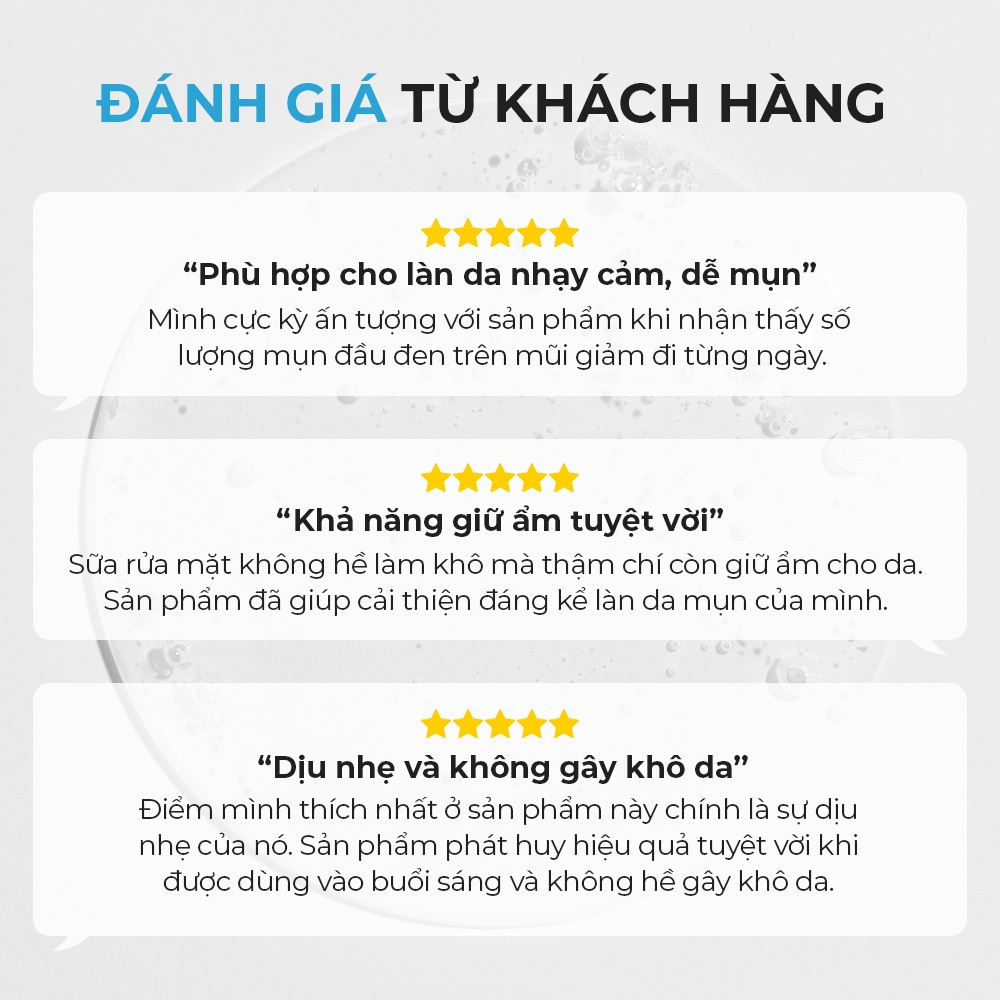}\hspace{0.02cm}
\includegraphics[width=0.12\linewidth]{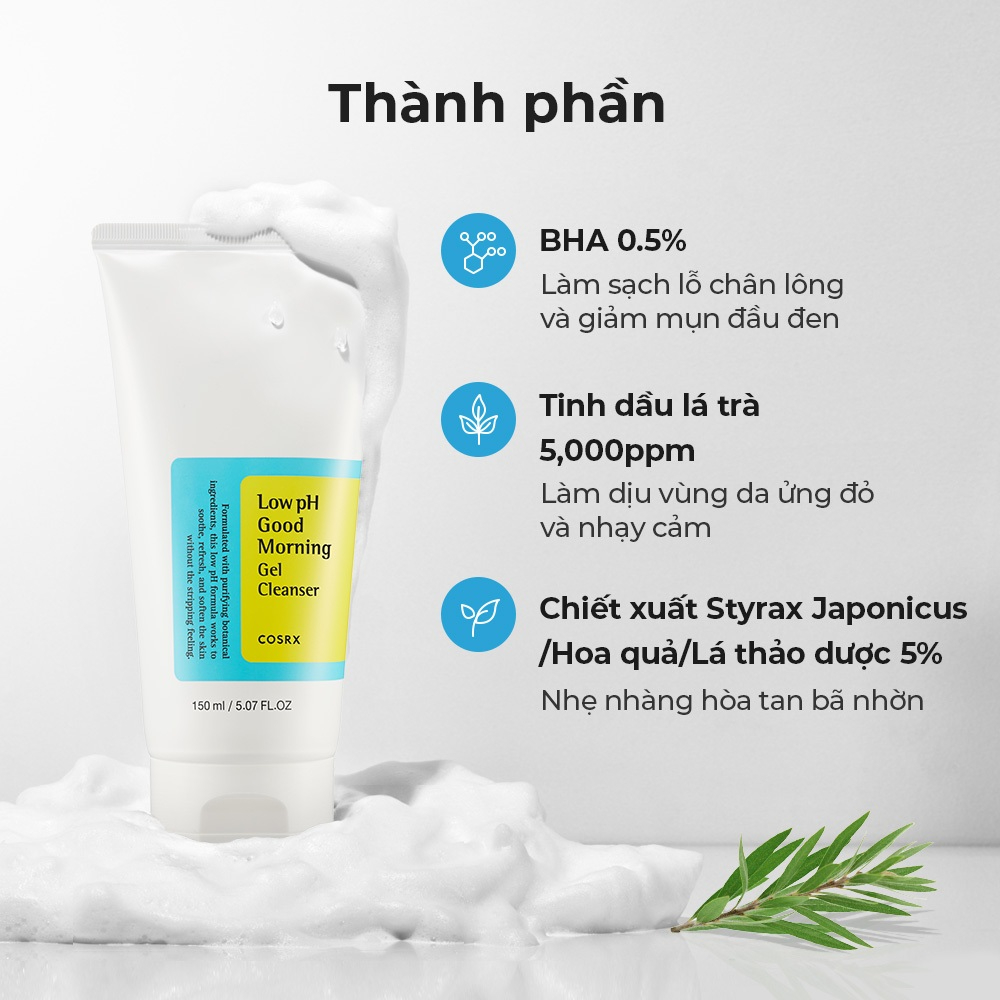}\hspace{0.02cm}
\includegraphics[width=0.12\linewidth]{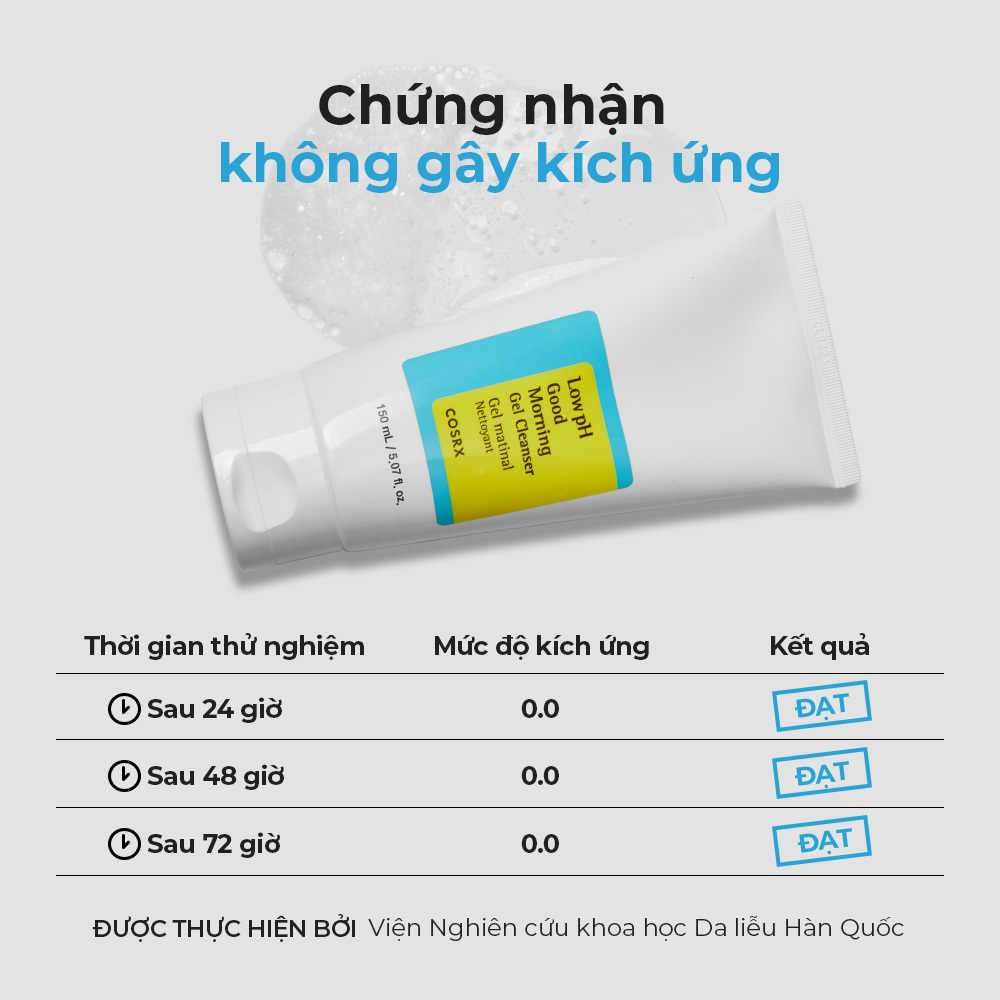}\hspace{0.02cm}
\includegraphics[width=0.12\linewidth]{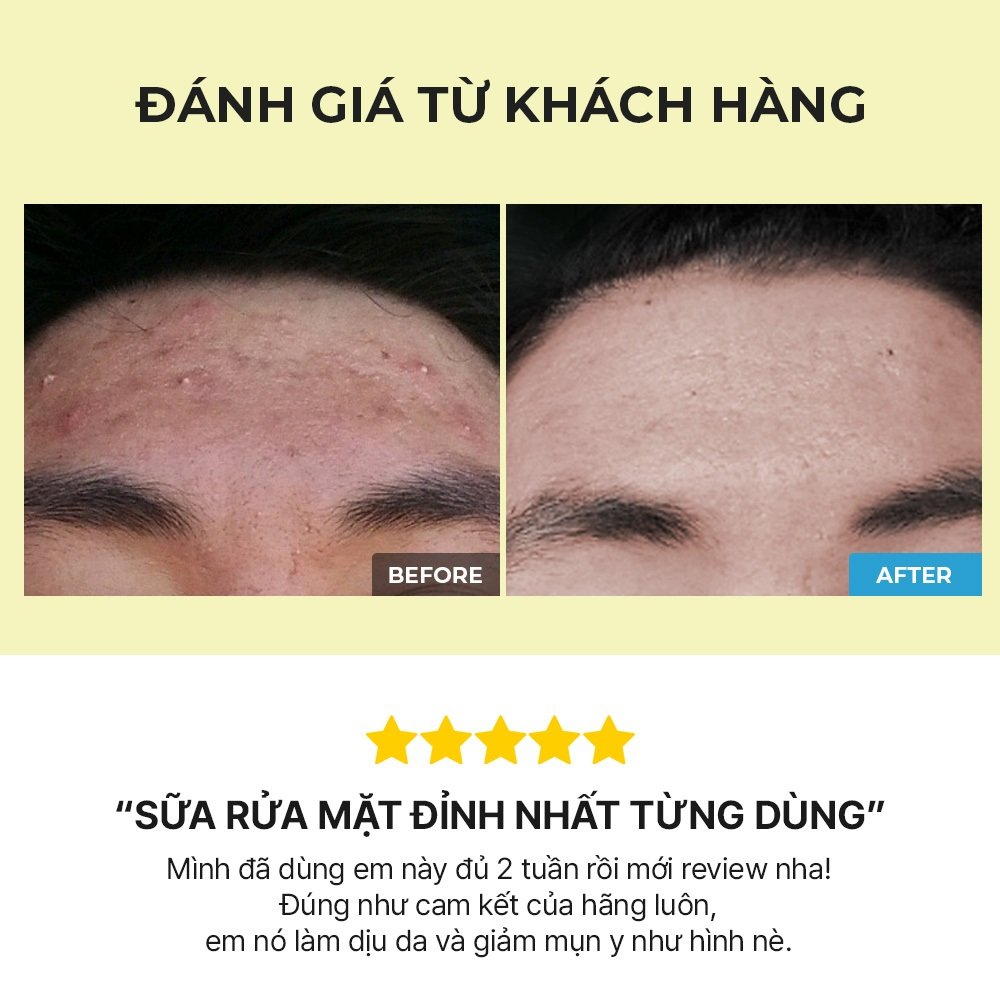}
\\
\midrule
\end{tabular}

\begin{tabular}{p{5cm} p{3.5cm} p{3.5cm}} 
\renewcommand{\arraystretch}{1.2}

\textbf{Review 1 \newline Helpfulness Score 4.0} & \textbf{Review 2 \newline Helpfulness Score 3.0} & \textbf{Review 3 \newline Helpfulness Score 2.0} \\
2 chai này là chai thứ 3 thứ 4 r đó \includegraphics[height=1em]{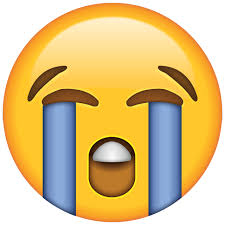} \includegraphics[height=1em]{Images/cry.jpg} \includegraphics[height=1em]{Images/cry.jpg} DÙNG MÊ ĐIÊN ĐẢO!!! ẻm không làm khô mặt mình sau khi rửa + dịu nhẹ nên mình hay dùng lắm!! Da mình cũng nhạy cảm nhma dùng tới chai thứ 3 thứ 4 là hiểu r đó, MUA LIỀN ĐIII \includegraphics[height=1em]{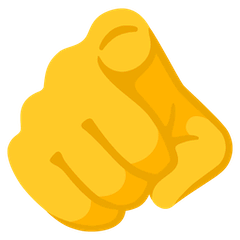} \includegraphics[height=1em]{Images/t.png} 
\newline
\includegraphics[width=0.15\linewidth]{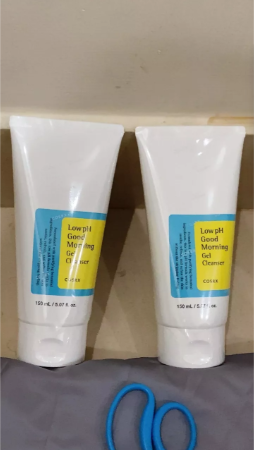} 
&
Nhãn ngoài hộp serum nổi bong bóng. Không biết có phải hàng chính hãng không 
\newline
\includegraphics[width=0.18\linewidth]{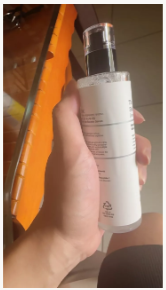} \hspace{0.001cm}
\includegraphics[width=0.18\linewidth]{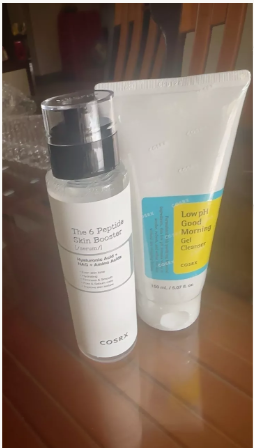}
&
Sữa rửa mặt thơm, rửa mặt sạch, shop giao hàng nhanh, đóng gói hàng cẩn thận
\newline
\includegraphics[width=0.18\linewidth]{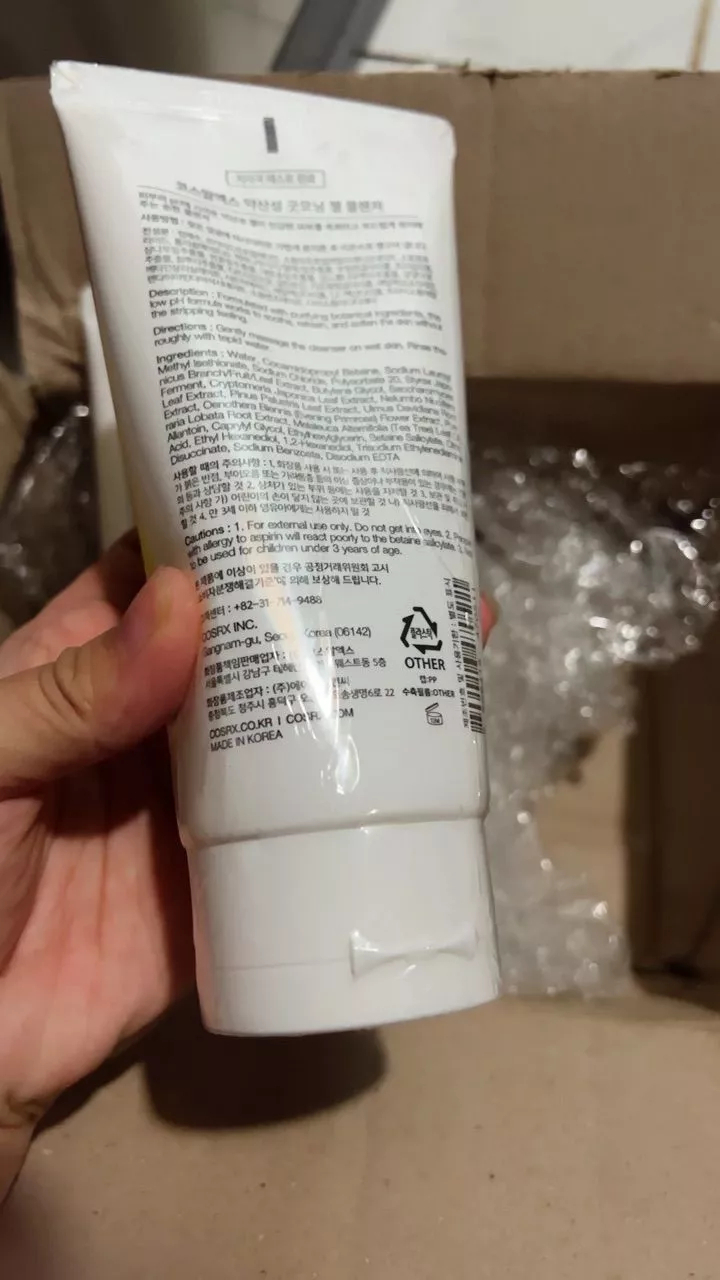} \hspace{0.001cm}
\includegraphics[width=0.18\linewidth]{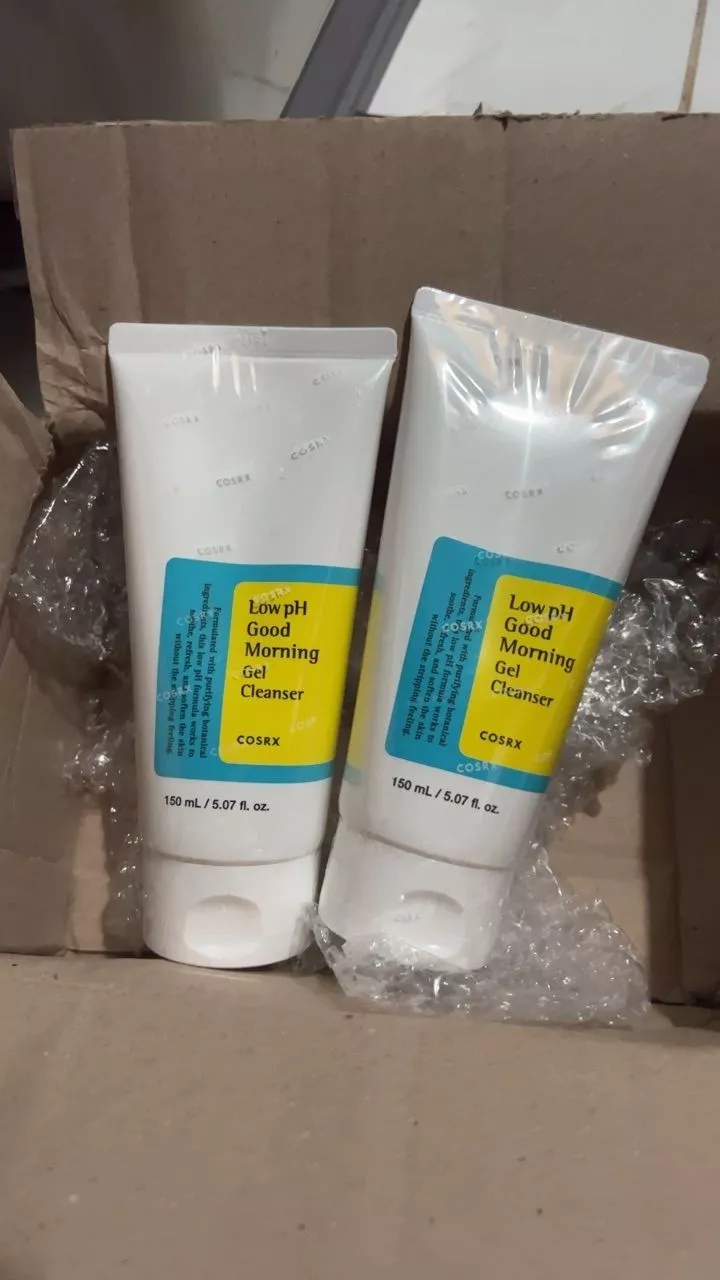} \\

\bottomrule
\end{tabular}

\caption{An illustrative example of the MRHP task and our ViMRHP dataset.}
\label{tab:ex}
\end{table}

Most existing studies focus on English, and research on MRHP in Vietnamese remains limited due to the scarcity of quality annotated datasets. This gap motivates our effort to propose a benchmark dataset for MRHP task in Vietnamese, formulated as a ranking task following the study of Liu et al. \cite{liu2021multi}, where reviews are ranked based on their helpfulness score, as detailed in \textbf{Table \ref{tab:ex}}.

However, data annotation for a high-quality dataset is often time-consuming and costly, requiring meticulous verification. To address this challenge, we leverage Large Language Models (LLMs) to assist in annotation, reducing manual effort while maintaining data quality. Despite their advantages in optimizing time and cost, LLMs still have limitations, requiring human verification to ensure accuracy and consistency. Therefore, we use a human-AI collaborative annotation framework that integrates LLMs with human verification. Additionally, we demonstrate the advantages and limitations of LLMs and human annotators in the ViMRHP dataset construction process through our evaluation metrics and experimental results. 

In summary, our main contributions are three-fold:

\begin{enumerate}
    \item \textbf{ViMRHP Dataset.} We introduce the ViMRHP dataset, a Vietnamese Multimodal Review Helpfulness Prediction dataset. To the best of our knowledge, there are no large-scale datasets for the MRHP task in Vietnamese.
    
    \item \textbf{Human-AI Collaborative Annotation.} Annotation process of our ViMRHP dataset implements Human-AI Collaborative Annotation framework via a two-step procedure: (1) AI annotation and (2) Human verification and refinement, ensuring time-efficiency, cost, and data quality.

    \item \textbf{Human-Verified versus AI-Annotated Data Quality.} We evaluate ViMRHP with baseline models to compare human-verified and AI-annotated data, highlighting differences in quality, consistency, and biases.

\end{enumerate}


%% file: Sections/2_RelatedWork.tex
\section{Related Work}

\subsection{Review Helpfulness Prediction}

Previous work, Review Helpfulness Prediction (RHP) has explored various tasks and approaches \cite{diaz2018modeling}. In the context of text reviews, the impact of structural, lexical, syntactic, semantic, and meta-data features on the helpfulness of user-generated reviews has been investigated\cite{kim2006automatically}, as well as semantic features\cite{yang2015semantic} for predicting review helpfulness using Amazon votes as ground truth and validating findings with the human-annotated label. Argument mining has also been studied separately, with the AM$^{2}$\cite{chen2022argument} dataset focusing on its role in determining the helpfulness of text review. Reviewer expertise and temporal dynamics \cite{nayeem2023role} have also been incorporated to enhance helpfulness prediction with dataset creation from user-generated reviews on TripAdvisor.

\begin{table}[!htbp]
    \centering
    \resizebox{\textwidth}{!}{%
    \begin{tabular}{lccccc}
        \toprule
        \textbf{Datasets} & \textbf{RHP \cite{nayeem2023role}} & \textbf{AM$^{2}$ \cite{chen2022argument}} & \textbf{Amazon-MRHP \cite{liu2021multi}} & \textbf{Lazada-MRHP \cite{liu2021multi}} & \textbf{ViMRHP (Ours)} \\ 
        \midrule
        \textbf{Annotation Method} & Vote Mapping & Human & Vote Mapping & Vote Mapping & Human-AI \\ 
        \textbf{Data Source} & TripAdvisor & Amazon & Amazon & Lazada & Shopee \\ 
        \textbf{Multimodal} & \xmark & \xmark & \cmark & \cmark & \cmark \\ 
        \textbf{Language} & English & English & English & Indonesian & Vietnamese \\ 
        \textbf{Task} & Multi-class CLS & Argument Mining & Ranking & Ranking & Ranking \\
        \hdashline 
        
        \textbf{No. of Reviews} & 161K & 878 & 1414K & 287K & 46K \\
        \textbf{No. of Products} & N/A & N/A & 59K & 21K & 2K \\
            \hdashline
         \multirow{4}{*}{\textbf{Domains}} & \multirow{4}{*}{\textit{Hotels}} & \multirow{4}{*}{\textit{Headphones}} & \textit{Clothing, Shoes \& Jewelry} & \textit{Clothing, Shoes \& Jewelry} & \textit{Fashion} \\ 
        & &  & \textit{Electronics} & \textit{Electronics} & \textit{Electronics} \\
        & &  & \textit{Home \& Kitchen} & \textit{Home \& Kitchen} & \textit{Home \& Lifestyle} \\
        & &  &  &  & \textit{Health \& Beauty} \\
        \bottomrule
    \end{tabular}%
    }
    \caption{Comparison of ViMRHP with notable Review Helpfulness Prediction (RHP) datasets. }
    \label{table:annotation_comparison}
\end{table}

In recent years, RHP has advanced by integrating textual and visual information\cite{liu2021multi, han2022sancl, nguyen2022adaptive, xu2024personalized}. Amazon-MRHP\cite{liu2021multi} and Lazada-MRHP\cite{liu2021multi} sourced from two major E-commerce platforms, Amazon and Lazada, serve as benchmarks for the MRHP task and have been utilized in MCR\cite{liu2021multi} Multi-perspective Coherence Reasoning is the first baseline for MRHP, SANCL\cite{han2022sancl} Selective Attention
and Natural Contrastive Learning model, which reduces GPU memory usage during training, Thong Nguyen et al.\cite{nguyen2022adaptive} propose Multimodal Contrastive Learning method achieved state-of-the-art results and PRR-LI\cite{xu2024personalized} a Large language model driven Personalized Review Recommendation
model based on Implicit dimension mining for MRHP. However, due to the limited availability of labeled data, most existing datasets \cite{nayeem2023role,liu2021multi} for this task defined ground truth by mapping helpful votes into helpfulness scores. We compare the proposed ViMRHP dataset with existing datasets in \textbf{Table \ref{table:annotation_comparison}}.

\subsection{Human-AI Collaborative Annotation Framework}
Reducing costs and ensuring high-quality annotations are two critical factors in NLP dataset construction. Recently, leveraging LLMs in annotation has become an optimized solution for time and cost efficiency \cite{wang2021want, ding2023gpt, tan2024large,he-etal-2024-annollm}. However, balancing time, cost, and data quality remains challenging for complex annotation tasks and datasets. Therefore, recent studies have explored the human-AI collaborative annotation framework to address this issue. Notable works such as CoAnnotating \cite{li-etal-2023-coannotating}, Wang et al.\cite{wang2024human}, and various datasets with diverse tasks have also adopted this framework, such as Value FULCRA \cite{yao2024value}, VIVA \cite{hu2024viva}. These works highlight the effectiveness of human-AI collaboration in enhancing annotation efficiency while maintaining quality.


%% file: Sections/3_AnnotationProcess.tex
\section{Human-AI Collaborative Annotation}\label{sec3_annotation_process}

\subsection{Overview}
\begin{figure}[!htpb]
    \centering
    \includegraphics[width=0.95\linewidth]{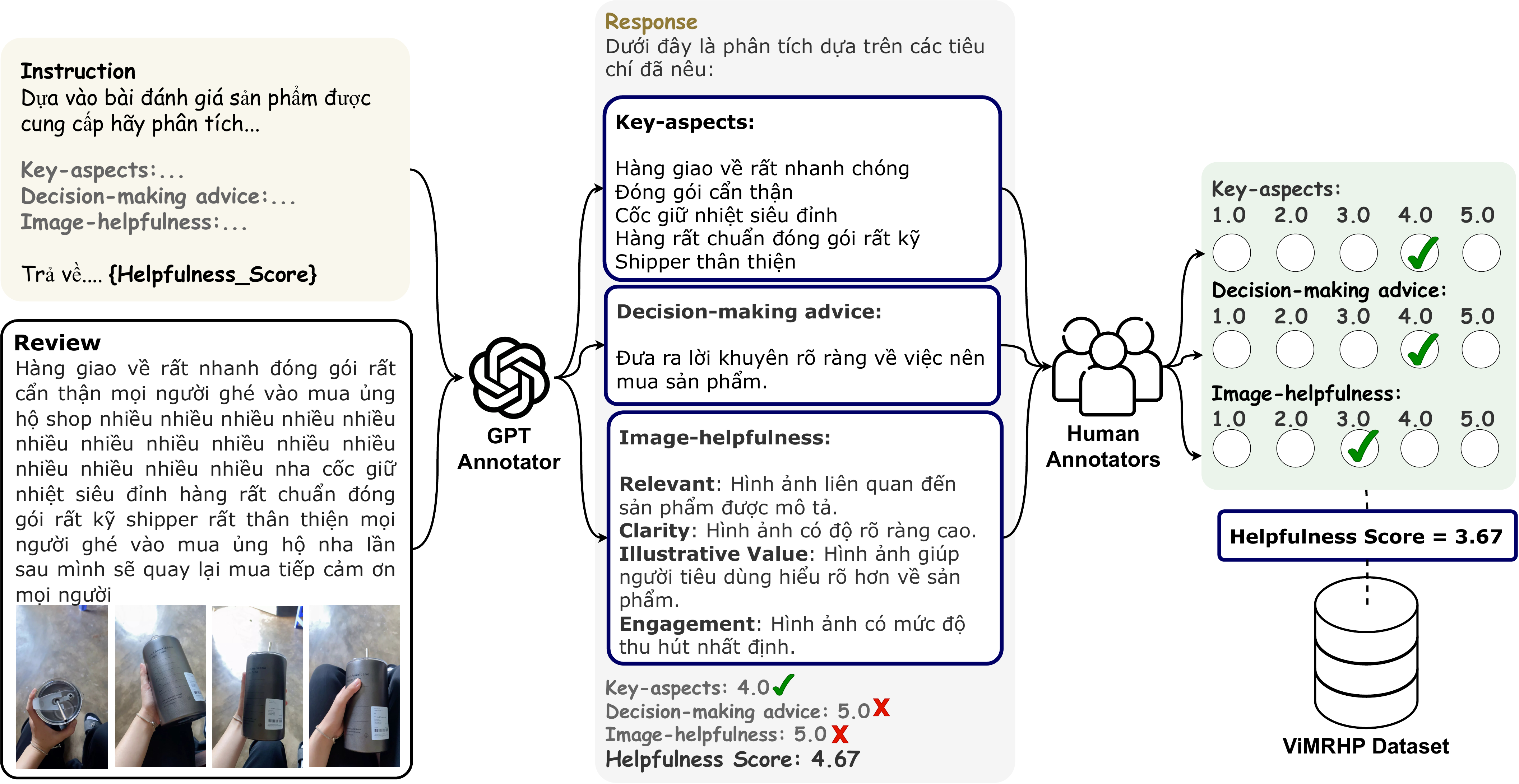} 
    \caption{ViMRHP benchmark dataset annotation overview. The Human-AI collaborative annotation framework workflow includes two steps: (1) AI Annotation $\rightarrow$ (2) Human Verification and Refinement. First, AI extracts the relevant context or gives a reason from the review based on the given instruction criteria and assigns a score. Then, human annotators verify and refine the final score to ensure data quality.
    \label{fig:pipeline}
}
\end{figure}
\clearpage

\textbf{Data Collection.} To construct the ViMRHP dataset, we collected product information and user-generated reviews, including both text and images, from the Shopee\footnote{\url{https://shopee.vn}} platform in Vietnam. For user-generated reviews, an average of \textit{21-24} reviews were collected per product, spanning the period from \textit{2019} to \textit{2024}. All user-related information was strictly removed before use to ensure privacy and compliance with ethical standards. 

\vspace{1mm}

\textbf{Task Annotation.} To ensure objectivity in evaluating review helpfulness, we define three key criteria for each sample in ViMRHP dataset: $\{(K_i, D_i, I_i)\}_{i=1}^{N}$, where $K_i$ is \textit{key aspects}, $D_i$ is \textit{decision-making advice}, and $I_i$ is \textit{image-helpfulness}, scoring on a scale from 1 to 5, where 1 is lowest and 5 is highest. The ground truth, referred to as the \textit{Helpfulness Score}, is computed as $H_i = \frac{1}{3} (K_i + D_i + I_i)$. The definition of each criterion and score is provided in \textbf{\S\ref{ss1}}. 

\vspace{1mm}

\textbf{Annotation Workflow.} 
We implemented a human-AI collaborative annotation framework for the ViMRHP dataset, which consists of two steps: Step 1 - AI Annotation (\textbf{\S\ref{ss2}}) and Step 2 - Human Verification and Refinement (\textbf{\S\ref{ss3}}). Details of our annotation process are illustrated in \textbf{Figure \ref{fig:pipeline}}. 

\subsection{Annotation Scheme and Guideline
}\label{ss1}
The annotation guideline for constructing the ViMRHP dataset ensures that each sample is individually labeled. This approach addresses the limitations of automated ground truth generation based on mapping the number of helpful votes, which may introduce bias, a common issue in previous studies, since not all reviews receive upvotes\cite{nayeem2023role}.  

\begin{table}[!ht]
    \centering
        \fontsize{8}{10}\selectfont

    \resizebox{\textwidth}{!}{
    \begin{tabular}{p{3.2cm} p{11.2cm}}
        \toprule
        \textbf{Criteria} & \textbf{Description} \\
        \midrule
        \multirow{5}{=}{\textit{Key-aspects:} The number of key aspects of the product mentioned in the user-generated review.
  
        } 
        & 1.0: Not mention any key aspects of the product. \\
        & 2.0: Mentions one key aspect.\\
        & 3.0: Mentions two key aspects. \\
        & 4.0: Mentions three key aspects. \\
        & 5.0: Mentions four or more key aspects. \\

        \hline
        \multirow{5}{=}{\textit{Decision-making advice:} The recommendation to purchase the product mentioned in the user-generated review.}        
        & 1.0: Describes an ambiguous experience without giving any purchase advice. \\
        & 2.0: Clearly describes the experience but does not provide purchase advice. \\
        & 3.0: Implicitly suggests whether the product is worth buying. \\
        & 4.0: Strongly implies whether the product is worth buying. \\
        & 5.0: Clearly recommends the product, specifying the target users or suitable situations. \\
    
        \hline
        
        \multirow{5}{=}{\textit{Image-helpfulness:} The level of usefulness of the product images provided by the user in the user-generated review. }
        & 1.0: Does not meet any criteria for Relevance, Clarity, Illustrative Value, or Engagement. \\
        & 2.0: Meets one criterion for Relevance, Clarity, Illustrative Value, or Engagement. \\
        & 3.0: Meets two criteria for Relevance, Clarity, Illustrative Value, or Engagement. \\
        & 4.0: Meets three criteria for Relevance, Clarity, Illustrative Value, or Engagement. \\
        & 5.0: Meets four criteria for Relevance, Clarity, Illustrative Value, and Engagement. \\

        \midrule
        \multicolumn{2}{c}{\textbf{Helpfulness Score = (Key-aspect + Decision-making advice + Image-helpfulness) / 3}}\\
    \bottomrule 
    \end{tabular}}
    \caption{Labeling Criteria for ViMRHP Dataset}
    \label{tab:labeling_criteria}
\end{table} 

Based on the work of Chua et al. \cite{chua2016helpfulness}, we establish a structured annotation framework using three key criteria: key aspects, decision-making advice, and image helpfulness. We assign a score to each criterion for each sample in our ViMRHP dataset. \textbf{Table \ref{tab:labeling_criteria}} presents a detailed labeling criteria, with the task design outlined below.

\begin{itemize}
    \item \textit{Key-aspects.}  Identify the context in which the review mentions aspects of the product (e.g., features, benefits, usage, durability, design, etc.) and assign a score accordingly, following the criteria in \textbf{Table \ref{tab:labeling_criteria}}.
    \item \textit{Decsion-making advice.} Identify whether the review provides purchasing recommendations (e.g., recommending to buy, not to buy, specifying target buyers, or advising against certain users, etc.) and assign a score based on its clarity and strength, following the criteria in \textbf{Table \ref{tab:labeling_criteria}}.
    \item \textit{Image-helpfulness.} Evaluate the helpfulness of user-uploaded images based on predefined criteria, including:
    \begin{itemize}
        \item Relevance: The image accurately represents the reviewed product.  
        \item Clarity: The image is clear and easily interpretable.
        \item Illustrative Value: The image effectively demonstrates key product features, benefits, or real-life usage.
        \item Engagement: The image captures user interest and enhances the review's informativeness.
    \end{itemize}
    Assign a score based on how well the image meets these criteria, following \textbf{Table \ref{tab:labeling_criteria}}.
\end{itemize}

An example annotation with detailed scoring for each criterion is shown in \textbf{Table \ref{tab:exm}}. For \textit{key aspects}, relevant contexts are highlighted and marked as (K1), (K2), (K3), (K4), and (K5). Similarly, for \textit{decision-making advice}, key statements are identified and marked as (D1). For \textit{image-helpfulness}, review images are assessed based on predefined criteria, ensuring a comprehensive evaluation.
\begin{table}[!htpb]
    \centering
    \fontsize{7.0}{9}
    \renewcommand{\arraystretch}{1.25} 
    \resizebox{\textwidth}{!}{
    \begin{tabular}{p{3.6cm} p{10.4cm}}
        \toprule

        \multirow{4}{*}{\includegraphics[width=0.285\linewidth]{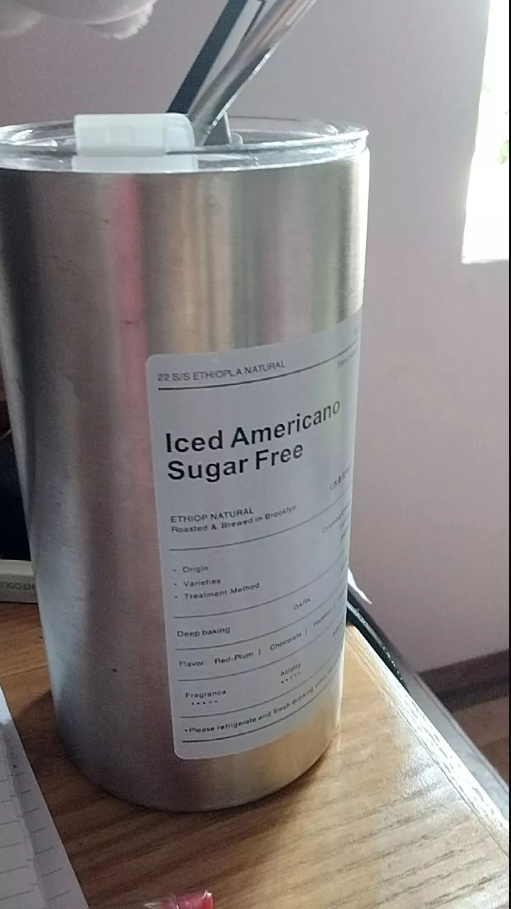}\hspace{0.011cm}
        \includegraphics[width=0.285\linewidth]{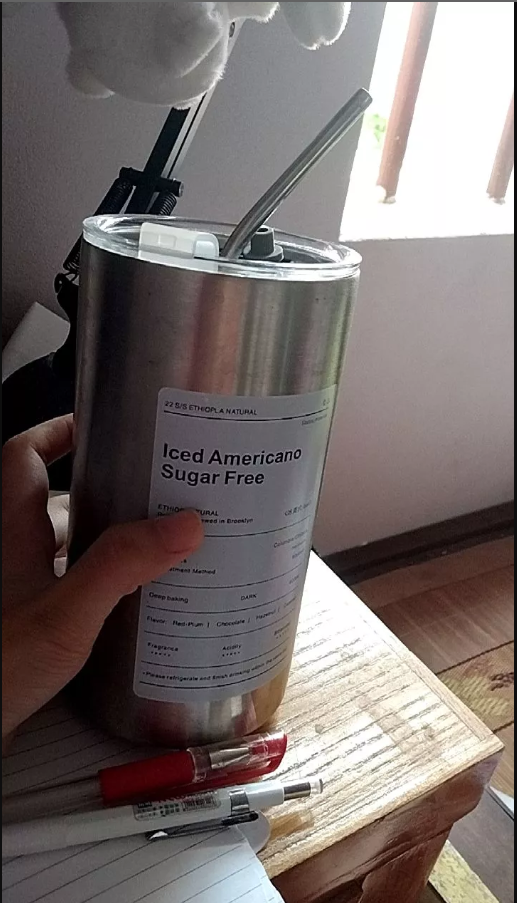}\hspace{0.011cm}
        \includegraphics[width=0.285\linewidth]{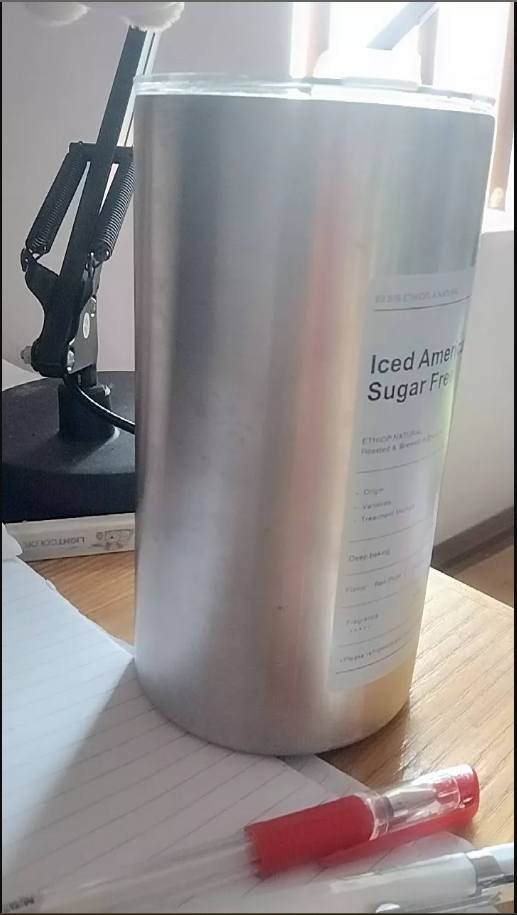}\hspace{0.011cm}
        }

        &     
        \colorbox{lightblue!40}{"ly khá là okla,  \underline{\textbf{k bị đọng nước ở ngoài}}\textsubscript{(K1)}, \underline{\textbf{thiết kế đơn giản thanh lịch 
         }}\textsubscript{(K2)}, }

        \colorbox{lightblue!40}{
        \underline{\textbf{k quá trơn}}\textsubscript{(K3)}, \underline{\textbf{nắp đậy k bị đổ}\textsubscript{(K4)}}, với giá vậy là rất rất rất tuỵt zời, 
                }

        \colorbox{lightblue!40}{
        còn về mức \underline{\textbf{độ giữ nhiệt thì mình thấy giữ được khoảng 6h }}\textsubscript{(K5)}, \underline{\textbf{rcm}}} 
        
        \colorbox{lightblue!40}{\underline{\textbf{nên muaa}}\textsubscript{(D1)}} \\

        \vspace{0.2mm}
       \underline{\textcolor{blue}{\textbf{Translation}}}:\colorbox{green!10}{{The cup is ok, 
        \underline{\textbf{doesn’t get condensation on the outside}\textsubscript{(K1)}}. \underline{\textbf{Design is simple elegant}}\textsubscript{(K2)}},} 
         \colorbox{green!10}{\underline{\textbf{not too slippery}}\textsubscript{(K3)} \underline{\textbf{The lid doesn’t leak}}\textsubscript{(K4)}. For this price, it’s absolutely great. As for heat retention,} 
         \colorbox{green!10}{\underline{\textbf{I found it keeps warm for about 6 hours.}} \textsubscript{(K5)}\underline{\textbf{ Recommend buying it}}\textsubscript{(D1)}.} \\

         \hdashline 

         \textbf{Key-aspects} 

         \textbf{Decision-making advice} 

         \textbf{Image-helpfulness} 
& 

         Mentions four or more key aspects: (K1),(K2),(K3),(K4),(K5) | \textcolor{blue}{\textbf{Score 5.0}}

         Strongly implies whether the product is worth buying: (D1) | \textcolor{blue}{\textbf{Score 4.0}}

         Meets two criteria. | \textcolor{blue}{\textbf{Score 3.0}} \\
         \hdashline

         \multicolumn{2}{l}{\textcolor{blue}{\textbf{Helpfulness Score: 4.0}}} \\

    \bottomrule 
    \end{tabular}}
    \caption{Example annotation for ViMRHP Dataset}
    \label{tab:exm}
\end{table}

\subsection{Step 1 - AI Annotation}\label{ss2}
We use LLM (\texttt{i.e., gpt-4o-mini version})\footnote{\url{https://platform.openai.com/docs/models/gpt-4o-mini}} \cite{gpt4omini} to automatically annotate approximately 46K review samples at a total cost of 150 - 170 USD,  significantly reducing expenses and time compared to manual annotation. Our annotation task involves two main challenges: (1) Extracting the review context that mentions aspects of the product for \textit{key-aspects}, explanation for \textit{decision-making advice} and \textit{image-helpfulness}. (2) Assign scores to each criterion. These are illustrated in the LLM response in \textbf{Figure \ref{fig:pipeline}} and detailed instructions are provided in Appendix \ref{app:instruction}.

\subsection{Step 2 - Human Verification and Refinement}\label{ss3}

In this step, to construct the ViMRHP dataset, we recruited three annotators, undergraduate students at our institution, with a Data Science background and Vietnamese proficiency. We paid 50 USD per annotator for modifying the entire dataset that AI-generated annotation in \textbf{Step 1}. Annotators were required to complete a training phase before verifying and refining AI-annotated data, during which they performed manual annotation on 100 samples. The inter-annotator agreement, measured using Fleiss's \(\kappa\), was assessed across three criteria: key aspects, decision-making advice, and image helpfulness. Their corresponding \(\kappa\) values were \(0.6341\), \(0.5944\), \(0.2107\). and 0.4484 for Helpfulness Score. Our labeling UI is detailed in the Appendix \ref{app:labelingUI}.


%% file: Sections/4_ViMRHP.tex
\section{ViMRHP Dataset}
\textbf{Dataset Statistics.} The statistical overview of the ViMRHP dataset is presented, details in \textbf{Table \ref{table:summary_statistics}}. The statistics comprehensively analyze the scale and features relevant to our multimodal ranking task, including Product Information (\textbf{P}) and Reviews (\textbf{R}). In addition, we provide details on the number of reviews per product (\textbf{R/P}), with our review list length ranging between 21 and 23 reviews per product. Furthermore, the number of images per review (\textbf{R}\textsubscript{img}\textbf{/R}) and the number of images per product (\textbf{P}\textsubscript{img}\textbf{/P}) provide valuable insights into multimodal aspects.

\begin{table}[!h]
    \centering
    \fontsize{8}{10}\selectfont
    \setlength{\tabcolsep}{2.2pt}
    \begin{tabular}{l ccc cc cc cc}
        \toprule
         & \multicolumn{3}{c}{avg.} & \multicolumn{2}{c}{avg. len} & \multicolumn{2}{c}{max. len} & \multicolumn{2}{c}{total} \\ 
        \cmidrule(lr){2-4} \cmidrule(lr){5-6} \cmidrule(lr){7-8} \cmidrule(lr){9-10}
         \textbf{Domain} & \textbf{R/P} & \textbf{P}\textsubscript{img}\textbf{/P} & \textbf{R}\textsubscript{img}\textbf{/R} & \textbf{P}\textsubscript{text} & \textbf{R}\textsubscript{text} & \textbf{P}\textsubscript{text} & \textbf{R}\textsubscript{text} & \textbf{P}\textsubscript{img} & \textbf{R}\textsubscript{img} \\ 
        \midrule
        Fashion          & 22.4 & 8.2 & 2.2 & 82.7 & 145.6 & 196 & 1782 & 4175 & 24857 \\ 
        Electronic       & 21.9 & 7.4 & 1.9 & 85.4 & 111.4 & 212 & 1292 & 3529 & 19289 \\ 
        Home \& Lifestyle & 22.5 & 7.6 & 2.0 & 86.6 & 118.2 & 157 & 2170 & 3391 & 20817 \\ 
        Health \& Beauty  & 22.9 & 7.7 & 2.4 & 79.6 & 129.2 & 206 & 2202 & 4794 & 34463 \\ 
        \bottomrule
    \end{tabular}
    \caption{Statistical overview of the ViMRHP dataset.}
    \label{table:summary_statistics}
\end{table}

Following the study of Liu et al. \cite{liu2021multi}, we split the ViMRHP dataset into \texttt{Train, Dev,} and \texttt{Test} sets with a ratio of 70:10:20, as detailed in \textbf{Table \ref{table:data_split}}, with the distribution of Products and Reviews splits also presented.
\begin{table}[!htbp]
    \centering
    \fontsize{7.8}{10}\selectfont
    \begin{minipage}{0.47\textwidth}
        \centering
        \begin{tabular}{l@{\hskip 0.8mm}c@{\hskip 1mm}c@{\hskip 1mm}c}
            \toprule
            \textbf{Domain} & \textbf{Train} & \textbf{Dev} & \textbf{Test} \\ \midrule
            Fashion          & 356/7812       & 50/1065      & 103/2132      \\ 
            Electronic       & 332/7274       & 47/1010      & 96/2101       \\ 
            Home\&Lifestyle  & 313/7153       & 44/1015      & 91/2057       \\ 
            Health\&Beauty   & 433/9956       & 61/1403      & 125/2832      \\ \bottomrule
        \end{tabular}
        \caption{\texttt{Train, Dev, Test} distribution across domains (Products / Reviews)}
        \label{table:data_split}
    \end{minipage}
    \hfill
    \begin{minipage}{0.52\textwidth}
        \centering
        \fontsize{7.8}{10}\selectfont
        \begin{tabular}{l@{\hskip 0.5mm}c@{\hskip 0.5mm}c@{\hskip 1mm}c}
            \toprule
            \textbf{Criteria} & \textbf{\%Agree} &  C$\kappa$ & \textbf{$\Delta$|H - A|} \\ 
            \midrule
            Key-aspects               & 40.34 & 22.96 & 81.29 \\ 
            Decision-making advice    & 52.93 & 34.65 & 64.00 \\ 
            Image-helpfulness         & 57.31 & 41.65 & 64.20 \\ 
            \textbf{Helpfulness Score} & \textbf{53.59} & \textbf{31.34} & \textbf{53.64} \\ 
            \bottomrule
        \end{tabular}
        \caption{Agreement evaluation between Human vs AI (\%)}
        \label{table:agreement_metrics}
    \end{minipage}
\end{table}

\textbf{In-depth Analysis - Human vs AI.} After verifying and refining the entire ViMRHP dataset, we compare the labels assigned by human annotators with AI-generated annotations using agreement metrics to assess consistency and accuracy, as detailed in \textbf{Table \ref{table:agreement_metrics}}. Specifically, we use three evaluation metrics: \%Agree - Human Agreement (Yes/No) \cite{chen2024mllm},  C$\kappa$ - Cohen's Kappa \cite{cohen1960coefficient}, and  $ \Delta|H - A|= \frac{1}{N} \sum_{i=1}^{N} | H_i - A_i |$, which quantifies the deviation in scores between human annotations and AI-generated labels across different criteria, providing a comprehensive measure of annotation reliability. Our analysis reveals varying agreement metrics, with Human Agreement ranging from 40.34\% to 57.31\% meaning human annotators manually refined approximately 50\% of the dataset to ensure data quality. Cohen’s Kappa scores between human vs AI for ground truth Helpfulness Score only 31.34\% indicating \textit{Fair Agreement}\cite{landis1977measurement}. The high $\Delta|H - A|$ deviation in \textit{key-aspects} 81.29\% suggests a significant gap in contextual understanding between AI and human annotators, limitations in accurately identifying contextual information.
\\

\begin{figure}[!ht]
    \centering
    \begin{subfigure}[b]{0.46\linewidth}
        \centering
        \includegraphics[width=\linewidth]{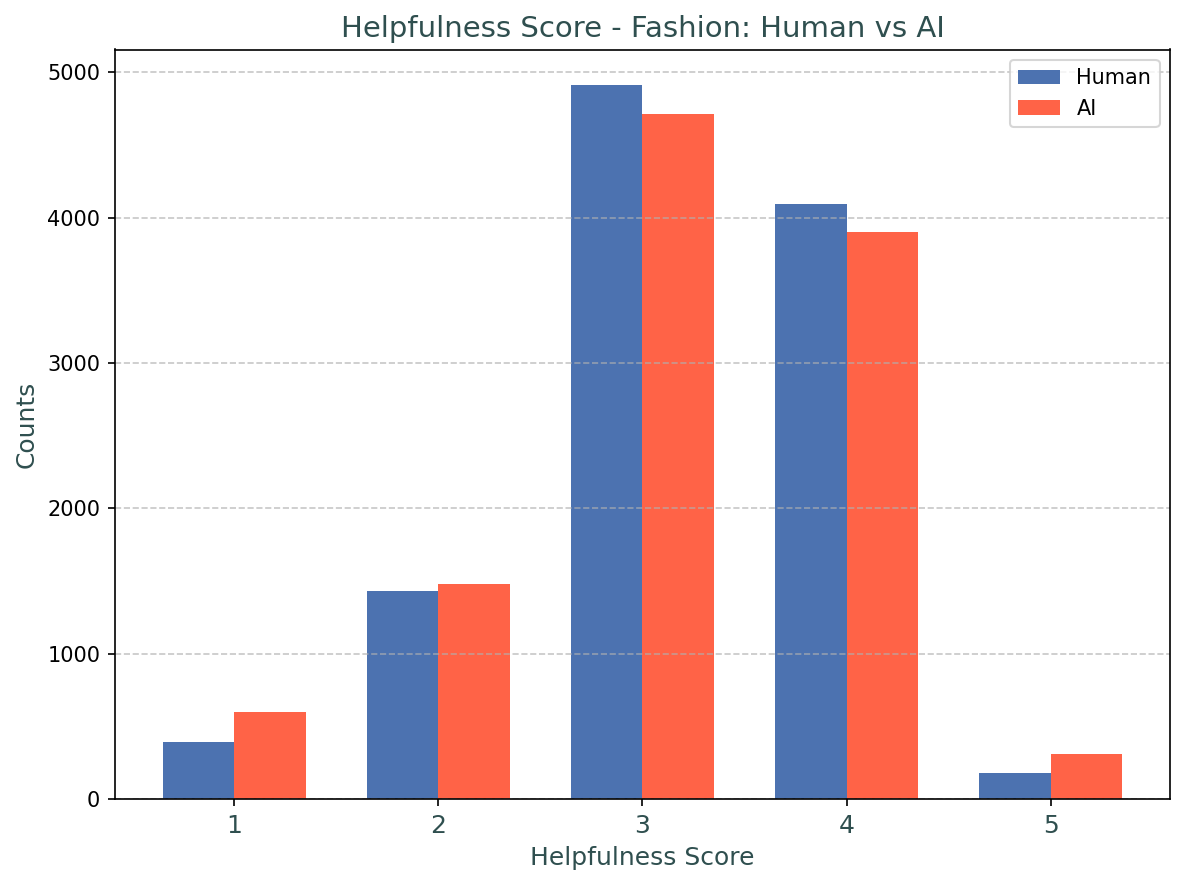}
        \caption{Fashion Category}
        \label{fig:score_fashion}
    \end{subfigure}
    \begin{subfigure}[b]{0.46\linewidth}
        \centering
        \includegraphics[width=\linewidth]{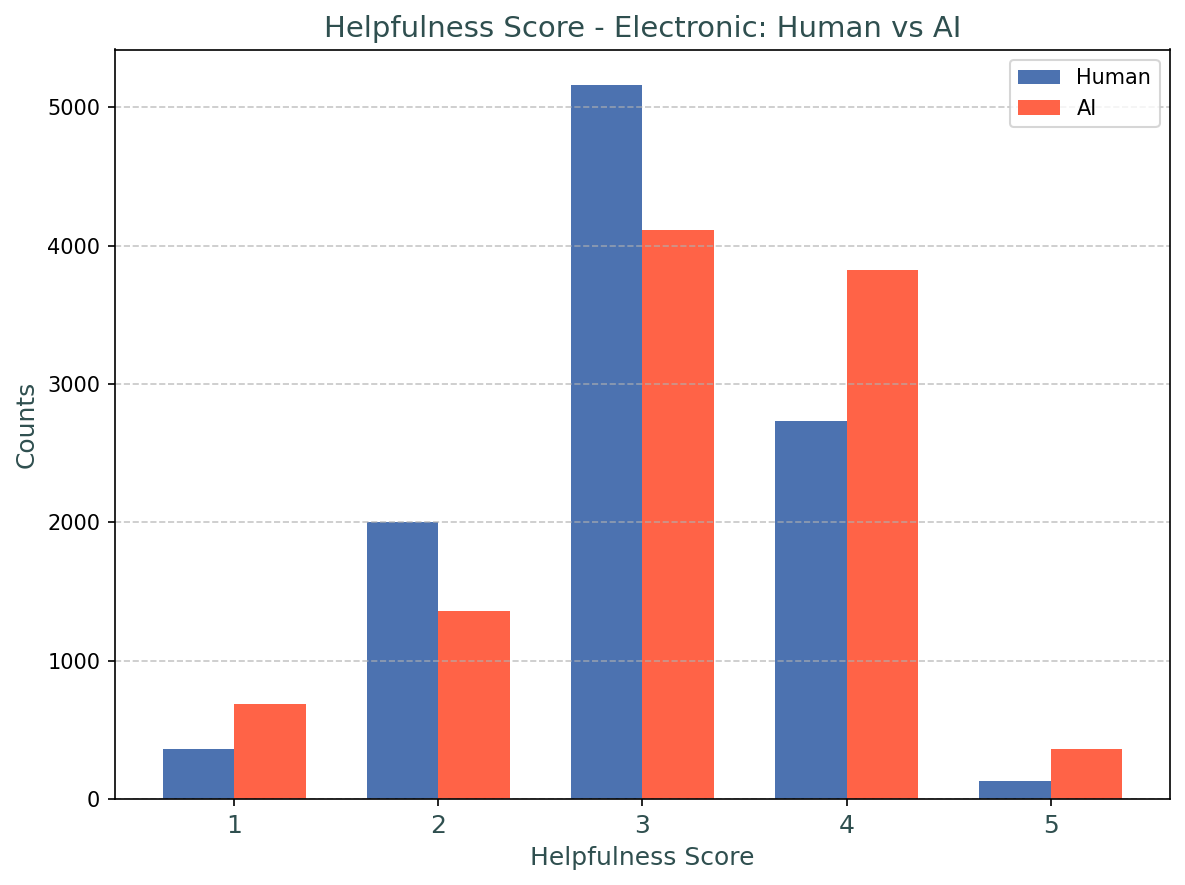}
        \caption{Electronic Category}
        \label{fig:score_electronic}
    \end{subfigure}
    \begin{subfigure}[b]{0.46\linewidth}
        \centering
        \includegraphics[width=\linewidth]{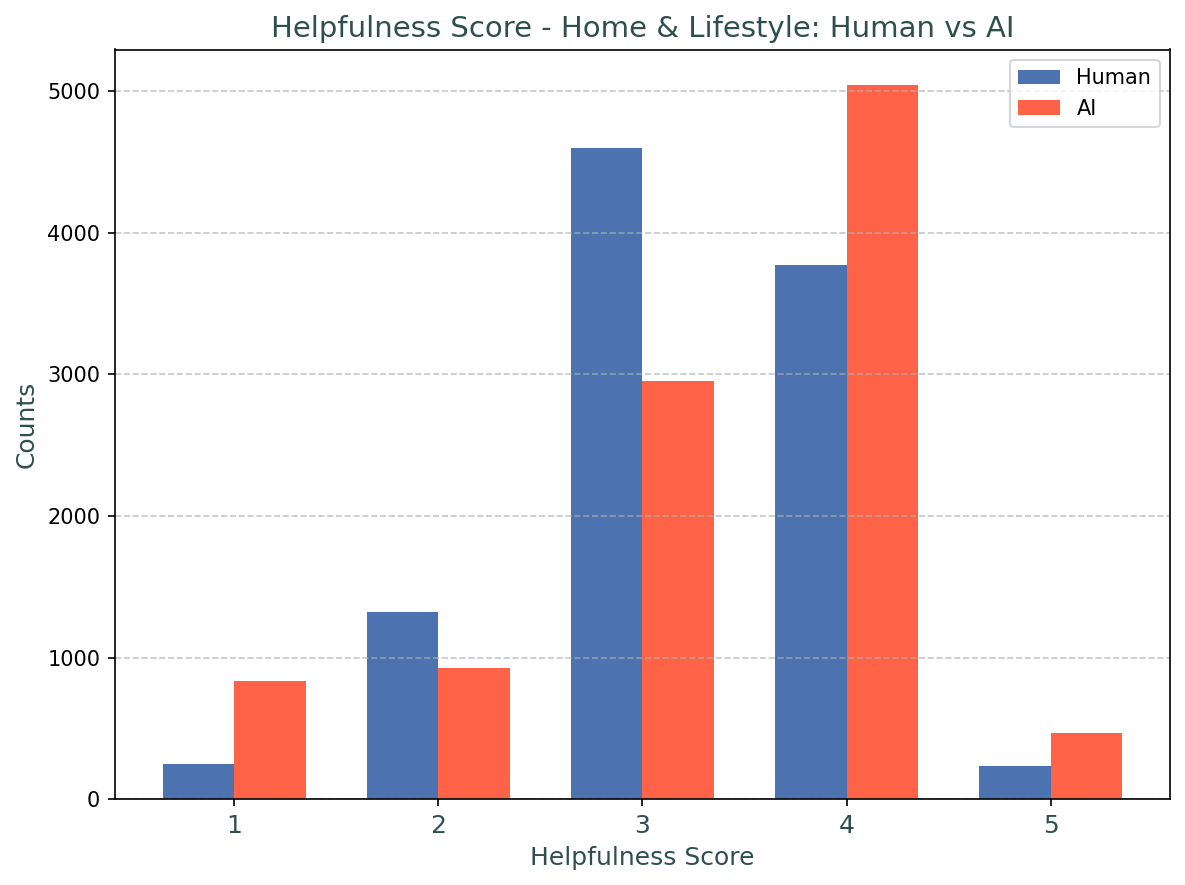}
        \caption{Home \& Lifestyle Category}
        \label{fig:score_home_lifestyle}
    \end{subfigure}
    \begin{subfigure}[b]{0.46\linewidth}
        \centering
        \includegraphics[width=\linewidth]{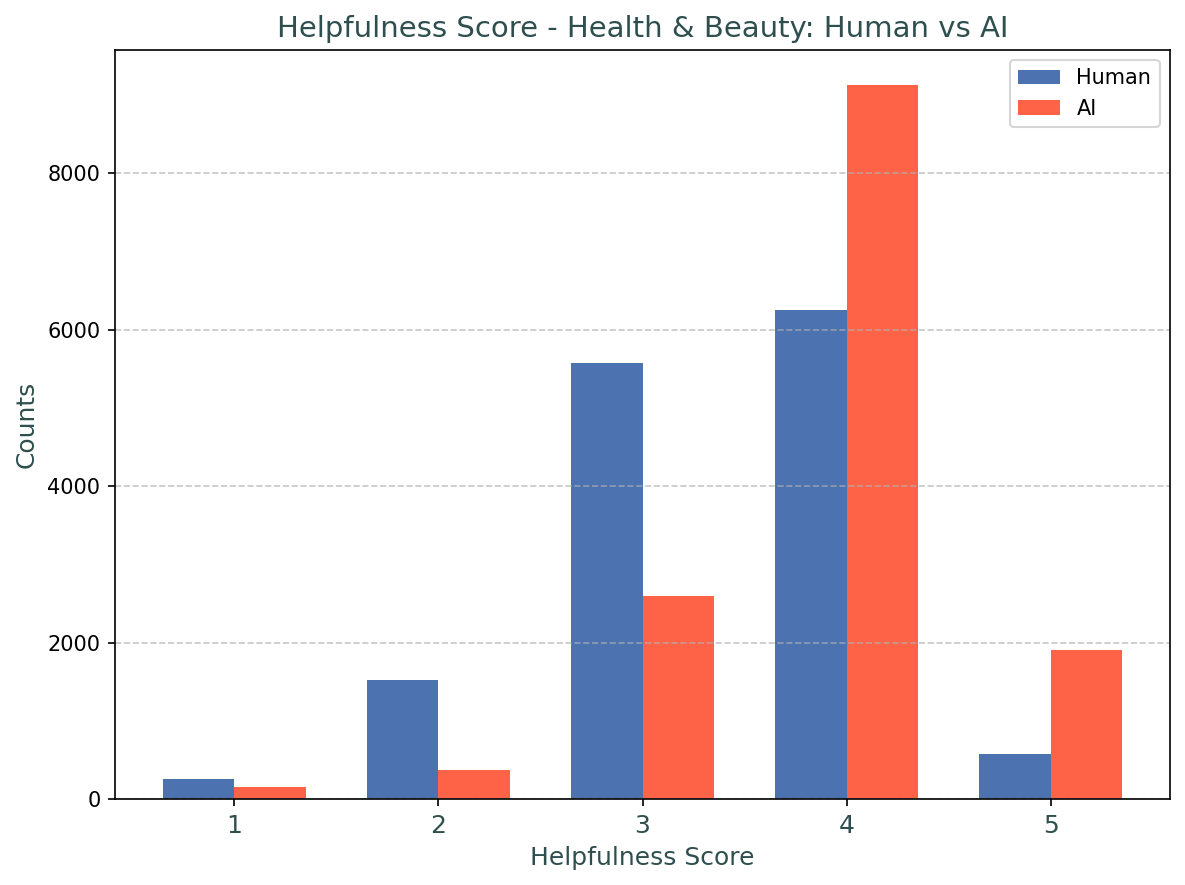}
        \caption{Health \& Beauty Category}
        \label{fig:score_health_beauty}
    \end{subfigure}
    \caption{Helpfulness Score distribution across categories between Human vs AI in ViMRHP dataset}
    \label{fig:four_images}
\end{figure}

\textbf{Distributions Analysis.} In ranking tasks, determining the distribution to select thresholds for experiments is crucial in evaluating a dataset. We analyze the distribution of common score ranges in each domain and highlight the scoring differences between human annotators and AI, detail in \textbf{Figure \ref{fig:four_images}} (\ref{fig:score_fashion}, \ref{fig:score_electronic}, \ref{fig:score_home_lifestyle}, \ref{fig:score_health_beauty}). The most common score range assigned is 3–4, reflecting a tendency toward neutral or safe reviews. This also serves as the basis for selecting our dataset evaluation threshold on the  average length of the review list per product (\textbf{R/P}) is 21–23, as detailed in \textbf{Table \ref{table:summary_statistics}}.\\
 \\



%% file: Sections/5_Experiments.tex
\section{Experimental Setup}

In our experiment, we implement various baseline models for our ViMRHP dataset, including text-only and multimodal approaches, to evaluate the contribution of each data modality to the review helpfulness prediction task.

\subsection{Baselines}

\subsubsection{Text-only}
\begin{itemize}[label=\textbullet]
    \item \textbf{BiMPM} \cite{wang2017bilateral} Bilateral Multi-Perspective Matching (BiMPM) uses BiLSTMs to encode and compare product information and reviews, capturing semantic relationships and aggregating relevance scores through a second BiLSTM.

    \item \textbf{Conv-KRMN} \cite{dai2018convolutional} is a neural ranking model designed to enhances ad-hoc search using CNNs to capture soft matches between query and document n-grams, which are mapped into a shared embedding space and processed through kernel-based pooling for relevance scoring.
    \item \textbf{DUET} \cite{mitra2017learning} is a neural ranking model that combines local interactions and distributed representations using two jointly trained deep networks to enhance lexical and semantic similarity in document ranking.
    \item \textbf{Match-Pyramid} \cite{pang2016text} is a CNN-based model for text matching that represents word-level similarities as a matching matrix. CNNs hierarchically extract complex patterns from this matrix, capturing key signals like n-gram and n-term matching.
\end{itemize}
\subsubsection{Multimodal}
\begin{itemize}[label=\textbullet]
    \item \textbf{MCR}\cite{liu2021multi} Multi-perspective coherence reasoning is a Multimodal Review Helpfulness Prediction baseline that integrates text and images from products and reviews. It includes two modules: one assessing intra- and inter-modal consistency between the product and review, and another ensuring coherence within the review by aligning textual and visual content.
\end{itemize}

\subsection{Evaluation Metrics}

Following the study by Liu et al. \cite{liu2021multi} on the Amazon-MRHP and Lazada-MRHP datasets, we adopt two evaluation metrics frequently used in recommendation: MAP (Mean Average Precision) and NDCG@K (Normalized Discounted Cumulative Gain), where \( K \in \{1,3,5\} \), to assess the ViMRHP dataset. These specific K values are chosen because users typically base their purchase decisions on the top few reviews, often only reading the first 1-5 reviews.

\subsection{Implementation Details}

Baseline models in our ViMRHP experiments are based on MatchZoo library \cite{guo2019matchzoo}, with all evaluations using a threshold of 3.0 for NDCG@K and MAP of MatchZoo \cite{guo2019matchzoo} metrics. All baselines, including text-only (BiMPM, Conv-KRMN, DUET, Match-Pyramid) and multimodal (MCR), word embedding layer are used with FastText, \texttt{batch\_size} 32, \texttt{learning\_rate} 0.001, Adam \texttt{optimizer} and executed on single GPU Tesla-T4-15GB. The text-only models are trained for 5 epochs per domain. The MCR model is trained for 20 epochs per domain. Before training, all product and review images are extracted with pre-trained Faster R-CNN\cite{anderson2018bottom} for RoI feature extraction.


%% file: Sections/6_Results.tex
\section{Results}
\subsection{Human-Verified versus AI-Annotated Data Quality.}\label{sec6-1}

We present the experimental results from our ViMRHP dataset in \textbf{Table \ref{table:exp_all_domains}}. The experiments were conducted using two types of annotated data: human-verified and AI-generated annotations. The results in \textbf{Table \ref{table:exp_all_domains}} show that multimodal approaches outperform text-only methods in the MCR baseline, especially with human-verified annotations in ViMRHP across Fashion, Electronics, Home \& Lifestyle, and Health \& Beauty, demonstrating the effectiveness of multimodal learning.  Additionally, leveraging LLMs can reduce costs and annotation time. However, human-verified annotations perform better across baseline models, underscoring the need for human verification to ensure data quality in complex annotation tasks.

\begin{table}[!ht]
    \centering
    \setlength{\emergencystretch}{4em}  
    \fontsize{4.6}{6.5}\selectfont
    \renewcommand{\arraystretch}{1.3} 
    \setlength{\tabcolsep}{1.5pt} 
    \begin{tabular}{lllcccccccc}
        \toprule
        \multirow{2}{*}{\textbf{Domain}} & \multirow{2}{*}{\textbf{Modality}} & \multirow{2}{*}{\textbf{Method}}  & \multicolumn{4}{c}{\textbf{Human Verification}} & \multicolumn{4}{c}{\textbf{AI Annotation}} \\ 
        \cmidrule(lr){4-7} \cmidrule(lr){8-11} 
         &  & & \textbf{N@1} & \textbf{N@3} & \textbf{N@5} & \textbf{MAP} & \textbf{N@1} & \textbf{N@3} & \textbf{N@5} & \textbf{MAP} \\
    
        \midrule
        \midrule
        
        \multirow{5}{*}{\textbf{Fashion}} 
        & \multirow{4}{*}{\textbf{Text-only}} & BiMPM & 64.80\textsubscript{\textcolor{blue}{1.51\textuparrow}} & 68.13\textsubscript{\textcolor{blue}{3.04\textuparrow}} & 71.15\textsubscript{\textcolor{blue}{0.70\textuparrow}} & 71.50\textsubscript{\textcolor{blue}{0.50\textuparrow}} & 
        63.29 & 65.09 & 70.45 & 71.00 \\

        && DUET & 
        
        34.73\textsubscript{\textcolor{blue}{0.87\textuparrow}} & 
        34.56\textsubscript{\textcolor{blue}{0.97\textuparrow}} & 38.03\textsubscript{\textcolor{blue}{0.87\textuparrow}} & 45.74\textsubscript{\textcolor{blue}{0.10\textuparrow}} & 

        33.86 & 33.59 & 37.16 & 45.64 \\
        && Conv-KRMN & 

        63.33\textsubscript{\textcolor{blue}{0.94\textuparrow}} & 63.51\textsubscript{\textcolor{blue}{0.68\textuparrow}} & 65.57\textsubscript{\textcolor{blue}{2.02\textuparrow}} & 67.86\textsubscript{\textcolor{blue}{1.48\textuparrow}} & 

        62.39 & 62.83 & 63.55 & 66.38 \\
        && Match-Pyramid & 
        63.89\textsubscript{\textcolor{blue}{9.74\textuparrow}} & 63.12\textsubscript{\textcolor{blue}{5.48\textuparrow}} & 65.01\textsubscript{\textcolor{blue}{5.49\textuparrow}} & 67.49\textsubscript{\textcolor{blue}{4.76\textuparrow}} & 
        
        54.15 & 57.64 & 59.52 & 62.73 \\

        & \textbf{Multimodal}
         & MCR & 
         \textbf{74.04}\textsubscript{\textcolor{blue}{5.36\textuparrow}} & \textbf{74.34}\textsubscript{\textcolor{blue}{2.48\textuparrow}} & \textbf{74.69}\textsubscript{\textcolor{blue}{2.16\textuparrow}} & \textbf{74.47}\textsubscript{\textcolor{blue}{1.27\textuparrow}} & 
         
         \textbf{68.68} & \textbf{71.86} & \textbf{72.53} & \textbf{73.20} \\
        
        \midrule
        \midrule
        
        \multirow{5}{*}{\textbf{Electronic}} 
        & \multirow{4}{*}{\textbf{Text-only}} & BiMPM & 
        
        62.33\textsubscript{\textcolor{blue}{16.09\textuparrow}}& 61.51\textsubscript{\textcolor{blue}{11.25\textuparrow}}& 63.10\textsubscript{\textcolor{blue}{10.54\textuparrow}}& 63.14\textsubscript{\textcolor{blue}{9.03\textuparrow}}& 
        
        46.24 & 50.26 & 52.56 & 54.11 \\

        & & DUET & 
        44.66\textsubscript{\textcolor{blue}{18.18\textuparrow}}& 51.59\textsubscript{\textcolor{blue}{25.97\textuparrow}}& 56.38\textsubscript{\textcolor{blue}{27.81\textuparrow}}& 57.78\textsubscript{\textcolor{blue}{22.32\textuparrow}}& 
        
        26.48 & 25.62 & 28.57 & 35.46 \\
        
        & & Conv-KRMN & 
        55.58\textsubscript{\textcolor{blue}{10.42\textuparrow}}& 56.13\textsubscript{\textcolor{blue}{10.71\textuparrow}}& 60.48\textsubscript{\textcolor{blue}{11.80\textuparrow}}& 61.51\textsubscript{\textcolor{blue}{10.52\textuparrow}}& 

        45.16 & 45.42 & 48.68 & 50.99 \\
        
        & & Match-Pyramid & 
        44.66\textsubscript{\textcolor{blue}{4.20\textuparrow}}& 49.53\textsubscript{\textcolor{blue}{7.62\textuparrow}}& 50.83\textsubscript{\textcolor{blue}{6.99\textuparrow}}& 53.27\textsubscript{\textcolor{blue}{5.61\textuparrow}}& 
        
        40.46 & 41.91 & 43.92 & 47.66 \\
        & \textbf{Multimodal} & MCR & 
        \textbf{69.15}\textsubscript{\textcolor{blue}{16.13\textuparrow}}& \textbf{65.25}\textsubscript{\textcolor{blue}{14.19\textuparrow}}& \textbf{66.16}\textsubscript{\textcolor{blue}{12.55\textuparrow}}& \textbf{66.00}\textsubscript{\textcolor{blue}{11.01\textuparrow}}& 

        \textbf{53.02} & \textbf{51.06} & \textbf{53.61} & \textbf{54.99} \\
        
        \midrule
        \midrule

        \multirow{5}{*}{\textbf{Home \& Lifestyle}} 
        & \multirow{4}{*}{\textbf{Text-only}} & BiMPM & 
        68.31\textsubscript{\textcolor{blue}{11.59\textuparrow}}& 73.65\textsubscript{\textcolor{blue}{13.17\textuparrow}}& 73.69\textsubscript{\textcolor{blue}{12.42\textuparrow}}& 75.18\textsubscript{\textcolor{blue}{10.08\textuparrow}}& 
        
        56.72 & 60.48 & 61.27 & 65.10 \\
        
        & & DUET & 
        35.34\textsubscript{\textcolor{blue}{4.59\textuparrow}}& 40.72\textsubscript{\textcolor{blue}{4.19\textuparrow}}& 43.30\textsubscript{\textcolor{blue}{4.19\textuparrow}}& 53.36\textsubscript{\textcolor{blue}{4.65\textuparrow}}& 
        
        38.75 & 36.53 & 39.11 & 48.71 \\

        & & Conv-KRMN & 
        50.76\textsubscript{\textcolor{blue}{6.66\textuparrow}}& 50.92\textsubscript{\textcolor{blue}{4.64\textuparrow}}& 54.32\textsubscript{\textcolor{blue}{5.44\textuparrow}}& 62.19\textsubscript{\textcolor{blue}{4.60\textuparrow}}& 
        
        44.10 & 46.28 & 48.88 & 57.59 \\

        & & Match-Pyramid & 
        62.78\textsubscript{\textcolor{blue}{8.22\textuparrow}}& 62.19\textsubscript{\textcolor{blue}{7.29\textuparrow}}& 62.34\textsubscript{\textcolor{blue}{7.34\textuparrow}}& 67.24\textsubscript{\textcolor{blue}{4.69\textuparrow}}& 
        
        54.56 & 57.88 & 58.00 & 62.55 \\
        & \textbf{Multimodal} & MCR & 

        \textbf{72.70}\textsubscript{\textcolor{blue}{10.98\textuparrow}}& \textbf{75.25}\textsubscript{\textcolor{blue}{13.07\textuparrow}}& \textbf{75.23}\textsubscript{\textcolor{blue}{11.63\textuparrow}}& \textbf{76.54}\textsubscript{\textcolor{blue}{9.72\textuparrow}}& 
        
        \textbf{61.72} & \textbf{62.18} & \textbf{63.60} & \textbf{66.82} \\

        \midrule
        \midrule
        
        \multirow{5}{*}{\textbf{Health \& Beauty}} 
        & \multirow{4}{*}{\textbf{Text-only}} & BiMPM & 

        69.96\textsubscript{\textcolor{blue}{6.76\textuparrow}}& 71.30\textsubscript{\textcolor{blue}{7.05\textuparrow}}& \textbf{72.60}\textsubscript{\textcolor{blue}{7.40\textuparrow}}& \textbf{77.36}\textsubscript{\textcolor{blue}{7.98\textuparrow}}& 
        
        63.20 & 64.59 & \textbf{65.20} & 70.38 \\
        & & DUET & 

        44.46\textsubscript{\textcolor{blue}{1.20\textuparrow}}& 47.88\textsubscript{\textcolor{blue}{6.00\textuparrow}}& 50.34\textsubscript{\textcolor{blue}{6.62\textuparrow}}& 60.84\textsubscript{\textcolor{blue}{4.60\textuparrow}}& 
        
        43.20 & 41.88 & 43.72 & 56.24 \\
        & & Conv-KRMN & 

        69.26\textsubscript{\textcolor{blue}{4.90\textuparrow}}& 69.70\textsubscript{\textcolor{blue}{5.58\textuparrow}}& 69.60\textsubscript{\textcolor{blue}{5.72\textuparrow}}& 74.05\textsubscript{\textcolor{blue}{4.35\textuparrow}}& 
        
        64.36 & 64.12 & 64.46 & 69.70 \\
        
        & & Match-Pyramid & 

        65.96\textsubscript{\textcolor{blue}{10.79\textuparrow}}& 65.88\textsubscript{\textcolor{blue}{16.13\textuparrow}}& 67.67\textsubscript{\textcolor{blue}{18.02\textuparrow}}& 73.31\textsubscript{\textcolor{blue}{7.55\textuparrow}}& 

        55.17 & 56.75 & 57.65 & 65.76 \\
        
        & \textbf{Multimodal} & MCR & 

        \textbf{71.59}\textsubscript{\textcolor{blue}{3.95\textuparrow}}& \textbf{73.22}\textsubscript{\textcolor{blue}{8.15\textuparrow}}& 72.47\textsubscript{\textcolor{blue}{7.63\textuparrow}}& 76.32\textsubscript{\textcolor{blue}{5.07\textuparrow}}& 
        
        \textbf{67.64} & \textbf{65.07} & 64.84 & \textbf{71.25} \\
        \bottomrule
    \end{tabular}
    \caption{Performance comparison for data quality in the ViMRHP dataset. Comparing human-verified data (\textbf{Human Verification}) with AI-generated annotations (\textbf{AI Annotation}). \textcolor{blue}{\textuparrow} denotes the percentage increase in performance.}
    \label{table:exp_all_domains}
\end{table}

\subsection{Cost, Time-Efficiency and Quality Comparison} 

We compare different annotation methods in \textbf{Table \ref{table:comparisionHuAI}}
for evaluating cost, time efficiency, and data quality. The annotation process of ViMRHP dataset collaborates with humans and LLMs (Section \textbf{\S\ref{sec3_annotation_process}}) costs around 300-320 USD. This cost achieves a balance, not as low as AI annotation, it remains significantly more affordable than the estimated cost of human annotation. With the support of LLMs, annotators achieved an annotation speed of 20-40 seconds per task, enabling the dataset to be completed within  3 weeks for approximately 46K multimodal reviews. 

\begin{table}[!h]
    \centering
    \fontsize{8.5}{10}\selectfont
    \setlength{\tabcolsep}{3.8pt} 
    \begin{tabular}{l@{\hskip 8mm}c@{\hskip 8mm}c@{\hskip 8mm}c}
        \toprule
         & \textbf{Human } & \textbf{AI } & \textbf{Human-AI } \\ \midrule
        \textbf{Cost} & 
        800 - 900 USD    & 150 - 170 USD      & 300 - 320 USD               \\  
        \textbf{Time-consume} & 
        2 - 3 months       & N/A               & 3 weeks                     \\ 
        \textbf{Efficiency} & 
        90 - 120s/Task     & 1 - 2s/Task       & 20 - 40s/Task               \\  
        \textbf{No. Annotators} & 
        9 - 12 annotators  & N/A               & 3 annotators                \\ 
        \textbf{Quality} & 
        \textbf{ \cmark} & \textbf{ \xmark} & \textbf{ \cmark}     \\ 
        \bottomrule
    \end{tabular}
    \caption{Comparison of annotation methods on the ViMRHP Dataset}
    \label{table:comparisionHuAI}
\end{table} 
 Moreover, based on the evaluation metrics between human and AI annotation in \textbf{Table \ref{table:agreement_metrics}} and the experimental results on ViMRHP dataset baselines in \textbf{Table \ref{table:exp_all_domains}}, we demonstrate that human-verified data achieves higher quality. To this end, human-AI collaboration is more efficient than traditional crowdsourced dataset creation in terms of data quality and reducing cost with time efficiency.

%% file: Sections/7_Conclusion.tex
\section{Conclusions}
This paper introduces ViMRHP, a Vietnamese dataset for MRHP tasks, covering four domains that reflect diverse user behaviors and data modalities. The ViMRHP dataset is a valuable resource for future research on MRHP tasks in Vietnamese, providing a quality dataset with human-verified annotations. Additionally, ViMRHP demonstrates a balance between cost, efficiency, and annotation quality with a Human-AI collaborative annotation framework to dataset construction. Furthermore, it highlights the limitations of LLMs in dataset creation, paving the way for more effective hybrid data annotation methods in the future.

%% file: Sections/Appendix.tex
\section{Labeling UI}\label{app:labelingUI}
 We utilize HumanSignal\footnote{Label Studio Enterprise Supported by HumanSignal Label Studio Academic Program}\cite{LabelStudio} as the labeling tool for the ViMRHP dataset shown in \textbf{Figure \ref{fig:ui}} below.
\begin{figure}[!htpb]{
    \centering
    \includegraphics[width=0.85\textwidth]{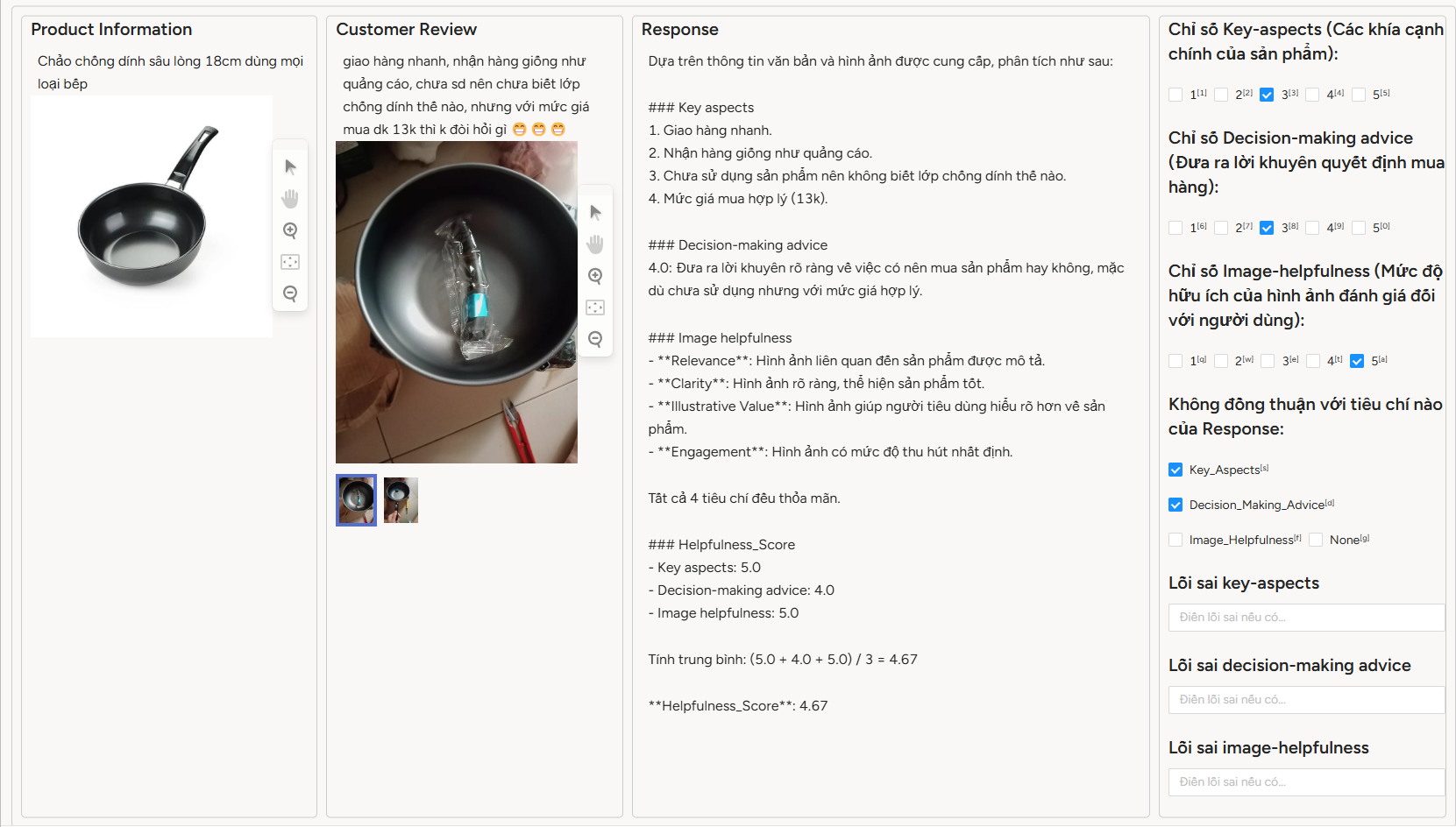} 
    \caption{Labeling UI for ViMRHP dataset}
    \label{fig:ui}
}
\end{figure}
\section{Instruction}\label{app:instruction}

\begin{table}[!ht]
    \centering
    \fontsize{7.2}{9}\selectfont
    \begin{tabular}{p{12cm}}
        \toprule

        
        \textcolor{darkred}{\textsc{Instruction (Vietnamese).}} Dựa vào thông tin bài đánh giá sản phẩm bao gồm văn bản và hình ảnh đã được cung cấp. Hãy
phân tích bài đánh giá theo các tiêu chí sau: \\

        \textbf{Key-aspects}: Đưa ra các khía cạnh chính của sản phẩm. \textbf{\{key\_aspects\}}
         \\  
        1.0 - Không đề cập đến khía cạnh cụ thể nào của sản phẩm.  
        
        2.0 - Đề cập đến một khía cạnh cụ thể của sản phẩm. 
        
        3.0 - Đề cập đến hai khía cạnh cụ thể của sản phẩm.
        
        4.0 - Đề cập đến ba khía cạnh cụ thể của sản phẩm. 
        
        5.0 - Đề cập đến nhiều hơn bốn khía cạnh cụ thể của sản phẩm. 
        
        \textbf{Decision-making advice}: Khuyến nghị mua hàng. \textbf{\{decision\_making\_advice\}} \\  
        1.0 - Mô tả trải nghiệm cá nhân mơ hồ, không đưa ra khuyến nghị. \\ 
        2.0 - Mô tả trải nghiệm cá nhân rõ ràng, nhưng không có khuyến nghị. \\ 
        3.0 - Ngầm đưa ra lời khuyên về việc có nên mua sản phẩm hay không.  \\  
        4.0 - Đưa ra lời khuyên rõ ràng về quyết định mua hàng. \\
        5.0 - Đưa ra lời khuyên cụ thể cho từng đối tượng khách hàng. \\
        \textbf{Image-helpfulness}: Mức độ hữu ích của hình ảnh sản phẩm dựa theo các tiêu chí.  \textbf{\{image\_helpfulness\}} \\   
        Mức độ liên quan (Relevance)..., Độ rõ ràng (Clarity)..., Giá trị minh họa (Illustrative Value)..., Tính thu hút (Engagement)... \\
         
        1.0 - Không thỏa mãn tiêu chí nào. \\  
        2.0 - Thỏa mãn một tiêu chí. \\  
        3.0 - Thỏa mãn hai tiêu chí. \\  
        4.0 - Thỏa mãn ba tiêu chí. \\  
        5.0 - Thỏa mãn cả bốn tiêu chí. \\  
        Trả về Helpfulness Score bằng trung bình điểm số ba tiêu chí \textbf{Key-aspects, Decision-making advice, Image-helpfulness}: 
        \textbf{\{Helpfulness\_Score\}} \\
        \hline
       \textcolor{blue}{\textsc{Instruction (English).}} Based on the provided product review, including both text and images, analyze the review according to the following criteria: \\

\textbf{Key-aspects:} Extract the main aspects of the product. \textbf{\{key\_aspects\}} \\  
1.0 - Does not mention any specific aspect of the product. \\  
2.0 - Mentions one specific aspect of the product. \\  
3.0 - Mentions two specific aspects of the product. \\  
4.0 - Mentions three specific aspects of the product. \\  
5.0 - Mentions more than four specific aspects of the product. \\  

\textbf{Decision-making advice:} Purchase recommendation. \textbf{\{decision\_making\_advice\}} \\  
1.0 - Describes an ambiguous experience without giving any purchase advice. \\
2.0 - Clearly describes the experience but does not provide purchase advice. \\
3.0 - Implicitly suggests whether the product is worth buying. \\
4.0 - Strongly implies whether the product is worth buying. \\
5.0 - Clearly recommends the product, specifying the target users or suitable situations. \\

\textbf{Image-helpfulness:} The helpfulness of product images based on the following criteria. \textbf{\{image\_helpfulness\}} \\  
Relevance..., Clarity..., Illustrative Value..., Engagement... \\  

1.0 - Does not meet any criteria. \\  
2.0 - Meets one criterion. \\  
3.0 - Meets two criteria. \\  
4.0 - Meets three criteria. \\  
5.0 - Meets all four criteria. \\  

Return \textbf{Helpfulness Score} is calculated as the average score of the three criteria: \textbf{Key-aspects, Decision-making advice, and Image-helpfulness.} \textbf{\{Helpfulness\_Score\}}\\

       
       
        \bottomrule
    \end{tabular}
    \caption{Prompt for AI-based Explanation Score}
    \label{tab:ai_evaluation}
\end{table}

\clearpage